\documentclass[journal]{IEEEtran} 

\IEEEoverridecommandlockouts
\usepackage{cite}
\usepackage{amsmath,amssymb,amsfonts}
\usepackage{algorithmic}
\usepackage{graphicx}
\usepackage{textcomp}
\usepackage{xcolor}
\usepackage{svg}
\usepackage{multirow}
\usepackage{caption}
\usepackage{subcaption}
\def\BibTeX{{\rm B\kern-.05em{\sc i\kern-.025em b}\kern-.08em
    T\kern-.1667em\lower.7ex\hbox{E}\kern-.125emX}}
\begin{document}

\title{SoftCTRL: Soft conservative KL-control of Transformer Reinforcement Learning for Autonomous Driving\\
\thanks{The computational resource for this work is supported by Innovation FabLab, Ho Chi Minh City University of Technology (HCMUT).}
}

\author{
    \IEEEauthorblockN{
    Minh Tri Huynh}
    , and \IEEEauthorblockN{
    Duc Dung Nguyen*, \textit{Member, IEEE}}
    \\
    \IEEEauthorblockA{\textit{AITech Lab., Ho Chi Minh City University of Technology (HCMUT), Vietnam}}\\
    \IEEEauthorblockA{\textit{Vietnam National University Ho Chi Minh City (VNUHCM), Vietnam}}\\
    \IEEEauthorblockA{\{tri.huynh\_tk15nbk, nddung\} @hcmut.edu.vn}
}

\maketitle

\begin{abstract}
In recent years, motion planning for urban self-driving cars (SDV) has become a popular problem due to its complex interaction of road components. To tackle this, many methods have relied on large-scale, human-sampled data processed through Imitation learning (IL). Although effective, IL alone cannot adequately handle safety and reliability concerns. Combining IL with Reinforcement learning (RL) by adding KL divergence between RL and IL policy to the RL loss can alleviate IL's weakness but suffer from over-conservation caused by covariate shift of IL. To address this limitation, we introduce a method that combines IL with RL using an implicit entropy-KL control that offers a simple way to reduce the over-conservation characteristic. In particular, we validate different challenging simulated urban scenarios from the unseen dataset, indicating that although IL can perform well in imitation tasks, our proposed method significantly improves robustness (over 17\% reduction in failures) and generates human-like driving behavior.
\end{abstract}


\section{Introduction} \label{INTRO}

Autonomous Driving Vehicles (AVs) have evolved rapidly in academic and industrial fields for decades. After the DARPA Grand Challenges in 2005-2007, many car manufacturers and startups rushed to develop the first level 4+ AVs.
In reality, making a safe, smooth, and deployable system is a major challenge for autonomous vehicles (AVs), especially under complex urban driving scenarios where the drivers must handle surrounding items' behaviors (such as bother vehicles, cyclists, pedestrians, other movable objects, and traffic light signals) before acting. This encourages learning-based methods, which are becoming more and more popular. One of these methods is an imitative learning-based approach proposed that allows the performance to scale with the amount of data available \cite{pomerleau1988alvinn, bojarski2016end, bansal2018chauffeurnet}. However, imitation learning (IL) is simple but suffers from covariate shift caused by the accumulation of errors \cite{ross2011reduction} and causal confusion \cite{de2019causal}.

Reinforcement learning (RL) has shown promising results in overcoming some limitations of traditional methods and IL by leveraging explicit reward functions to increase safety awareness \cite{kendall2019learning, kothari2021drivergym, zhang2021end}. In addition, RL policies can also form causal relationships between states, actions, and outcomes by modeling sequential decision-making problems. Unfortunately, many modern RL algorithms heavily rely on reward design, an open challenge in AV.

With a human-designed reward design, driving policies using RL may be technically safe but unnatural. Many methods \cite{schaal1996learning, kober2010imitation, Nair2017Overcoming, Vecerik2017Leveraging, Rajeswaran2017learning, fujimoto2021minimalist, kumar2020conservative, wu2019behavior, Siegel2020Keep, nair2020accelerating, lu2023imitation} resolve this by combining IL and RL since they offer complementary strengths. RL enhances safety and robustness, especially in long-tail scenarios while IL improves realism and lessens the load on reward design. Therefore, combining IL and RL is crucial to enhance realism with large-scale human driving datasets and safety with causal relationship modeling. 

However, combining IL and RL is challenging. Utilizing IL (teacher) policy to constrain RL (student) policy may lead to the over-conservation problem caused by covariate shift of IL. This issue happens when IL trained with logged datasets guide RL students in online training process, where exploration is possible. Accumulated single-step errors can cause planned states to diverge from the dataset distribution, resulting in a compounding error issue. Therefore, this constraint force causes the RL student to be too conservative with the estimated expert policies to prevent the model from searching for a global solution.

To address these problems, we propose a \textbf{Soft} conservative KL-\textbf{C}on\textbf{T}rol of \textbf{RL} model (SoftCTRL) for autonomous driving, a novel approach for off-policy RL. First, we use a pretrained Transformer IL model to constrain RL updates with simple rewards. During the training step, the RL policy is implicitly regularized by the KL divergence from this prior model and the entropy of its actions. The entropy term encourages the model to generate more diverse trajectories. This combats the over-conservation issues when retaining the proximity to the distribution of the realistic IL model by KL control. Our experiments show that a simple reward combining imitation with collision is sufficient for our proposed method. Also, our new approach beat the pretrained IL model by a large margin in overall tasks, demonstrating that our approach is not affected by the covariate shift of IL.

The main contributions of our work are summarized as follows:

\begin{itemize}
    \item We proposed a novel off-policy RL method using implicit entropy-KL control to incorporate both IL and RL advantages. This approach substantially improves the learning speed, balances realism and safety in driving, and combats the over-conservation issue, reducing failure by 17\% over IL-alone and 41\% over RL-alone.
    \item We further study the complementarity role of entropy and KL divergence to balance between the unnatural (diversity) and conservation (realism) of the model.
\end{itemize}

\section{Related Works} \label{RW}

\subsection{Learning-based approaches in autonomous driving}

Many data-driven SDV systems have recently developed, exploiting breakthroughs in deep learning methods. We briefly summarize the signatures of different learning-based algorithms for motion forecasting and planning in SDV.  Imitation learning (IL) and inverse reinforcement learning (IRL) are scalable ML approaches that utilize large expert demonstration data. These methods aim to directly mimic expert behavior or recover the underlying expert costs.  Early research on simple open-loop behavior cloning dates back to 1989 when the ALVINN system \cite{pomerleau1988alvinn} employed a 3-layer neural network to perform lane keeping using front camera images. Although open-loop IL shows significant progress \cite{bojarski2016end, salzmann2020trajectron++, chitta2022transfuser, renz2022plant, hu2023uniAD}, it suffers from covariate shift issue \cite{ross2011reduction} (which can be alleviated with perturbation \cite{bansal2018chauffeurnet} or closed-loop training \cite{bhattacharyya2022rail, ho2016gail,baram2016mgail, scheel2022urban}). Furthermore, IL techniques do not inherently possess explicit knowledge of essential driving components, such as collision avoidance. Finally, a majority of human data consists of trivial scenarios, such as remaining stationary or driving in a straight line at a nearly constant speed; therefore IL becomes an unreliable and not robust method in complex and long-tail scenarios like urban driving. 
RL, on the other hand, can tackle this problem by leveraging the Markov decision process (MDP) model to learn the long-term consequence of its action. Moreover, RL allows the policy to learn from explicit and non-differentiable reward signals with closed-loop training and has been applied to specific tasks such as lane-keeping \cite{kendall2019learning}, intersection traversal \cite{isele2018navigating}, and lane changing \cite{wang2018reinforcement} or general tasks like driving in urban area like \cite{kothari2021drivergym}.

RL and other closed-loop methods require a simulator environment to train and validate. In recent years, many heuristic-based simulators CARLA \cite{dosovitskiy2017carla} and SUMO \cite{lopez2018sumo} show a significant improvement in realism. However, there is a huge gap between the hand-coded simulator and real-world behaviors. Highway-env \cite{leurent2018highenv} and TORCS \cite{wymann2000torcs} are OpenAI Gym\cite{brockman2016openaigym} compatible simplified, compact settings for autonomous driving but lack essential semantic aspects such as traffic lights, a detailed evaluation procedure, and expert data. Data-driven simulators are intended to address this issue by utilizing real-world traffic logs to simulate. VISTA \cite{amini2020vista} developed a photorealistic simulator for training a complete RL policy. \cite{henaff2019modelpredictive} created a bird's-eye picture of congested freeway traffic. Recently,  TrafficSim \cite{suo2021trafficsim}, SimNet \cite{bergamini2021simnet}, and DriverGym \cite{kothari2021drivergym} powered by large-scale datasets \cite{houston2021one, chang2019argoverse}, created mid-representation closed-loop simulators and showed their utility in training and verifying ML planners. In our experiments, we train and evaluate our methods in closed-loop on real-world data with other agents following logs.

\subsection{Encoding Map and Interaction}
One popular class of map representation is rendering inputs as a multi-channel rasterized top-down image, which has been used by many visual-based approaches \cite{bansal2018chauffeurnet, cui2019multimodal, chai2019multipath, chang2019argoverse}. This approach captures relationships between elements on a map via convolution neural networks (CNNs). Although CNN-based approaches show simplicity, they suffer from high rendering time and high-dimensional input space, which may cost a large network, training, and inference time. Moreover,  rasterized images lose explicit relations between agent-agent and agent-map, which requires more data to recover those relationships \cite{huang2022multi}. More recently, a notably compact vector map representation \cite{gao2020vectornet} has captured a collection of static and dynamic elements around the ego, which can significantly reduce computational cost. VectorNet \cite{gao2020vectornet} tokenizes the lanes on the map and the historical trajectories of the agent as a set of polylines. The key idea is to create a global fully connected interaction graph that captures the relationships between scene elements, which are processed by a graph neural network (GNN). PlanT \cite{renz2022plant}, wayformer \cite{nayakanti2023wayformer}, MTR \cite{shi2022mtr}, MTR++\cite{shi2024mtr++} utilize transformer encoder layers to extract features from different vehicles and routes, modeling interpretable relations between entities to predict actions. Urban Driver \cite{scheel2022urban} presents a PointNet-based version of VectorNet \cite{gao2020vectornet}, as well as provides closed-loop training and evaluation. In this work, we use the Urban Driver \cite{scheel2022urban} architecture as a Transformer backbone for downstream RL motion planning tasks.

\subsection{Combining imitation and reinforcement learning}

Combining RL and IL is, in fact, not new. Prior works attempted to combine IL and RL (IL$+$RL), by pre-training RL policies with demonstrations \cite{schaal1996learning}, incorporating demonstration data into off-policy RL algorithms \cite{kober2010imitation}, and combining IL and RL loss functions into a single algorithm \cite{Nair2017Overcoming, lu2023imitation}. BC-SAC \cite{lu2023imitation} conducted a combined IL and RL approach utilizing large amounts of real-world urban human driving data and a simple reward function to address challenges in autonomous driving. Also, DDPGfD \cite{Vecerik2017Leveraging}, and DAPG \cite{Rajeswaran2017learning} combine demonstrations with RL by including a margin loss which encourages the expert actions to have higher Q-values than all other actions. Offline RL approaches, such as TD3+BC \cite{fujimoto2021minimalist} and CQL \cite{kumar2020conservative} combine RL objectives with IL ones to regularize Q-learning updates and avoid overestimating out-of-distribution values. RL with Kullback-Leibler (KL) regularization is a particularly successful approach. In KL-regularized RL, the standard RL objective is augmented by a KL term that penalizes the dissimilarity between the RL policy and a BC reference policy derived from expert demonstrations. Recent research has shown that KL-regularized objectives in offline RL, such as BRAC \cite{wu2019behavior} and ABM \cite{Siegel2020Keep}, improve sample efficiency and solve challenging environments previously unsolved by standard deep RL algorithms. However, these algorithms struggle with online fine-tuning due to covariate shifts, making KL constraints overly conservative. To mitigate over-conservativeness, BC-SAC \cite{lu2023imitation} introduces a weighted IL loss directly into the policy improvement step, without using pre-trained IL, allowing modeling the mixture of online RL and offline IL data for urban driving. AWAC \cite{nair2020accelerating}, an implicit KL-regularized RL approach, applies advantage-weighted updates that allow deviations from the IL if beneficial. Our approach offers a different solution to the problem by introducing an implicit entropy-KL regularization method, with tunable entropy and KL parameters, to balance policy diversity and conservativeness.

\section{Background and Notations} \label{BG}
\subsection{Imitation Learning} \label{BG:IL}
Imitation learning optimizes a policy by imitating an expert using large human data. This method assumes our expert policy is optimal, denoted as $\pi_{e}$, which generates a dataset of trajectories $\mathcal{D} = \{s_0, a_0, s_1, a_1,... s_N, a_N\}$ by interacting with the environment. The IL learner's goal is to train a policy $\pi^{\text{IL}}_{\theta}$ that imitates the expert. This problem is similar to supervised learning in that the input is expert states and the output label is extracted from expert action data. Behavioral cloning (BC) trains the policy to minimize the negative log-likelihood objectives of the expert dataset, the optimal parameters are:

\begin{equation}
\theta^{*} = \arg\min_{\theta} - \mathbb{E}_{s,a\sim \mathcal{D}} \left[\text{log} (\pi^{\text{IL}}_{\theta}(a \mid s)\right]
\end{equation}
To alleviate covariate shift,  ChauffeurNet  \cite{bansal2018chauffeurnet} proposed a state perturbation during training BC that forces to learn how to recover from drifting, leading to learning robust driving policies Alternatively, closed-loop approaches include adversarial IL (GAIL \cite{ho2016gail}, MGAIL\cite{baram2016mgail}), which aim to align the state-action visitation distribution between the policy and the expert through a discriminator. On the other hand, methods such as offline policy gradient with BPTT \cite{scheel2022urban} use a differentiable simulator to compute policy gradients based on multi-step past trajectories directly. Unlike adversarial IL, which aligns distributions through an adversarial game, the offline policy gradient method updates the policy directly by backpropagating through time. Both approaches can help resolve the covariate shift issue that affects open-loop IL, though they use fundamentally different mechanisms.
\subsection{Reinforcement learning} \label{BG:RL}
Reinforcement learning uses MDPs (Markov Decision Processes) as a formal way to define the interaction between a learning agent and its environment. Let $\Delta_X$ be the set of probability distributions over a finite set $X$ and $Y^X$ the set of applications from $X$ to the set $Y$. A discounted MDP \cite{sutton2018reinforcement} contains a 5-tuple $(\mathcal{S}, \mathcal{A},  P, r, \gamma)$, where $\mathcal{S}$ and $\mathcal{A}$ are the state and action spaces, $ P \in \Delta_{\cal S,\cal A}^{\cal S}$ are the transition function, $r \in \mathbb{R}^{\cal S \times \cal A}$ is the reward function and $\gamma$ is a discount factor. A policy $\pi\in \Delta_{\cal A}^{\cal S}$ defines either an action (deterministic policy) or a distribution over actions (stochastic policy) given a state. We will also use $\rho_\pi(s_t)$ and  $\rho_\pi(s_t, a_t)$ to denote the state and state-action marginals of the trajectory distribution induced by a policy $\pi(\cdot\mid s_t ).$ The state-action value functions are defined as:
\begin{equation}
q_\pi = \mathbb{E}_\pi[\sum_{t=0}^{T} \gamma^{t}r(S_t, A_t) \mid S_0 = s, A_0 = a]
\end{equation}

The goal is to find the optimal policy $\pi^{\ast}$, which results in the highest expected sum of discounted rewards. In standard MDP settings, a deterministic greedy policy satisfies:
\begin{equation}
\label{eq:pi_objective}
       \pi^{\ast}= \arg \max_{\pi}  \mathbb{E}_\pi\left[\sum_{t=0}^{T} \gamma^{t}r_{t}\right]
\end{equation}
To do so, $q$-learning methods (e.g. Deep Q-learning (DQN) \cite{mnih2013dqn}) learn to estimate state-action value function $q$ approximated by an Q-network $q_{\phi}$, with weights copied regularly to a target network $q_{\bar{\phi}}$, through stochastic gradient descent (SGD) on the mean-square Bellman error (MSBE) loss $
J_q(\phi) = \mathbb{E}_{\mathcal D}\left[\frac{1}{2}
    \big( q_{\phi}(s_t,a_t) - \hat{q}_{\text{dqn}}(r_t,s_{t+1}) \big)^2
    \right]$, where the target $\hat{q}_{\text{dqn}}$:
\begin{equation}
\label{eq:q_objective}
\hat{q}_{\text{dqn}}(r_t, s_{t+1}) = r_t + \gamma \mathbb{E}_{s' \sim \rho_\pi(s, a)}[max_{a'}q_{\bar{\phi}}(s', a')]
\end{equation}

Maximum entropy RL (e.g. Soft Q-learning (SQL) \cite{haarnoja2017SQL}, Soft Actor Critic (SAC) \cite{haarnoja2018SAC}) is an entropy-regularized deviation of the MDP approach in which the reward is augmented with an entropy term. The optimal policy aims to maximize both cumulative reward and its entropy $\mathcal{H}(\pi) = - \mathbb{E}_{a \sim \pi}\big[\log (\pi(\cdot \mid s_t))\big]$ at each visited state:
\begin{equation}
\label{eq:pi_soft_objective}
\begin{aligned}
\pi^{*}_{\text{MaxEnt}} 
&= \arg \max_{\pi}  \mathbb{E}_\pi\left[\sum_{t=0}^{T} \gamma^{t} r_{t}\right] + \tau \mathcal{H}(\pi) \\ 
&= \text{softmax}(\frac{1}{\tau}q^{*}_\text{soft}(s_t, a_t)) \\
\end{aligned}
\end{equation}
where $\tau$ is the temperature of the entropy term. Note that when $\tau \rightarrow 0$ then \textit{softmax} becomes regular \textit{argmax} mentioned in \eqref{eq:pi_objective}, and we retrieve DQN.  Because entropy-regularized RL avoids taking a hard max over noisy estimated Q during training, it leads to less overestimation of future Q value and improves the stability of temporal-difference (TD) updates. In the $q$-value estimation step, SQL modifies the regression target as follows:

\begin{multline}
\label{eq:sac-critic-target}
\hat{q}_{\text{MaxEnt}}(r_t,s_{t+1})= r_t + \gamma \mathbb{E}_{a'\sim\pi_{\bar{\phi}}} \big[ q_{\bar{\phi}}(s_{t+1}, a') \\ - \tau \log \pi_{\bar{\phi}}(a'|s_{t+1}) \big] 
\end{multline} 
with $\pi_{\bar{\phi}} = \text{softmax}(\frac{q_{\bar{\phi}}}{\tau})$. The $- \tau \log\pi_{\bar{\phi}}(a'|s_{t+1})$ term is analogous to entropy regularization scaled by $\tau$. To extend to SAC \cite{haarnoja2018SAC}, we simply replace the discrete action $a'$ with the squashed Gaussian distribution. SAC also uses function approximations for estimating policy (parameterized by $\theta$) instead of $q$-value. The actor policy network $\pi_{\theta}$ outputs the mean $\mu$ and deviation $\sigma$ of each possible action which is parameterized as a Diagonal Gaussian distribution $N(\mu, \sigma^2)$. In the actor update, we assume that the $q$-value has been estimated. The policy acts to maximize the state value function $v_\pi$  by SGD: $J_\pi(\theta) = \arg \max_{\pi_\theta\in\Delta_{\cal A}^{\cal S}}v_{\pi_\theta}$ , where $v_{\pi_\theta}$ can be expanded out into:
\begin{equation}
\label{eq:sac-actor-update}
v_{\pi_\theta}  =  \mathbb{E}_{a'\sim\pi_\theta}\big[q_\phi -\tau \log \pi_{\theta}(a' \mid s)\big]
\end{equation}

\section{Soft Conservative KL-control for Reinforcement Learning (SoftCTRL)}

\subsection{Model Architecture}
\begin{figure}[!t]
    \begin{center}
    \includegraphics[width=\linewidth]{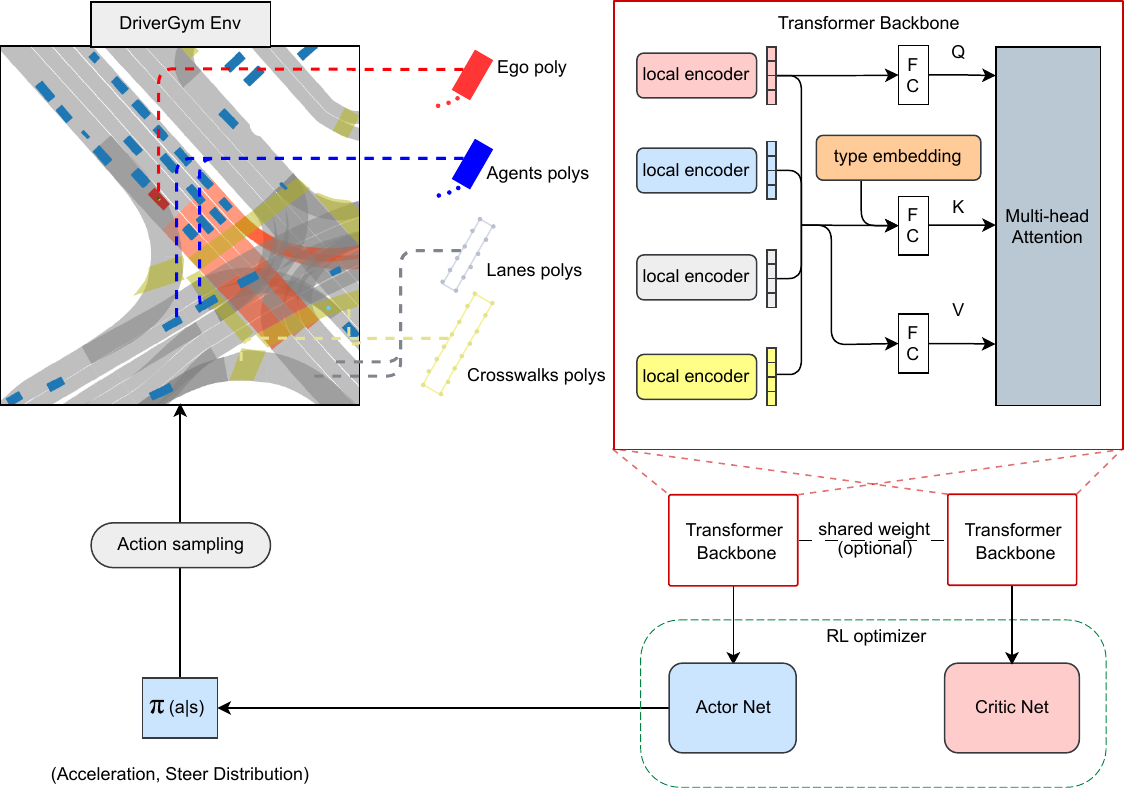}
    \end{center}
    \caption[Overview of our Transformer-enhanced Reinforcement learning framework.]{Overview of our transformer-enhanced Reinforcement learning framework.}
    \label{fig:rl_pipeline}
\end{figure}

Our RL model is shown in Fig.~\ref{fig:rl_pipeline}, which comprises 2 main parts: the backbone feature extractor network and the actor-critic network. We employ a Transformer architecture in \cite{scheel2022urban} for feature extraction that encodes efficient vectorized input into a high-dimensional feature space. The resulting feature is then fed into the latter network, called the actor-critic net. To boost RL training, we leverage the pretrained Transformer-based behavioral cloning model in\cite{scheel2022urban} to initialize our RL's feature extractors. 
Those feature extractor parameters are either trainable or initialized from IL and freezing. The shared weight between the backbones of the actor and critic net is optional. After that, our RL agent samples actions from the current policy distribution to interact with the environment. This training loop drives the agent to seek long-term and maximum overall rewards to achieve an optimal solution. The detail of the transformer backbone is in section~\ref{sec:appendix_model_details}.

\subsection{Soft Conservation with generalized Entropy-KL control}

Our approach designs a soft conservative control for combining the IL-RL approach to incorporate IL and RL's complementary strengths and combat over-conservation problems. Conventional KL-regularized RL approaches minimize a KL term $\mathbb{D}_{\mathrm{KL}}\big(\pi(\cdot \mid s_t) || \pi_0(\cdot \mid s_t)\big)$ while training, resulting in the over-conservation issue. Alternatively, entropy-based RL approaches that maximize the entropy of the policy distribution $\mathcal{H}(\pi(\cdot \mid s_t)$ exhibit to mitigate the overestimation of the future estimated Q value, which is similar to the mentioned over-conservation problem. Intuitively, merging KL and entropy terms offers the advantage of softening the over-conservation tendency inherent in KL divergence loss. One simple approach is directly adding KL loss to the reward function of entropy-regularized RL algorithms but it is hard to control the trade-off between entropy and KL. In this work, we extend the entropy-regularized RL to a generalized entropy-KL RL using a reparameterization trick. Consequently, we maximize the following combined objective function during the RL training process:
\begin{multline}
\label{eq:pi_KL_objective}
       \pi^{\ast}_{\text{EntKL}}= \arg \max_{\pi}  \mathbb{E}_\pi\left[\sum_{t=0}^{T} \gamma^{t}r_{t}\right]  + \tau \mathcal{H}(\pi(\cdot \mid s_t) \\ - \alpha \mathbb{D}_{\mathrm{KL}}\big(\pi(\cdot \mid s_t) || \pi_0(\cdot \mid s_t)\big) 
\end{multline}

Instead of directly optimizing entropy-regularized RL with an additional KL reward to achieve both entropy and KL regularization, we employ the idea of Munchausen RL \cite{vieillard2020munchausen}, which implicitly maximizes its entropy and conservative update policy with KL divergence. To obtain implicit entropy-KL control, we simply modify the regression target of the policy evaluation step by adding the scaled log policy of reference behavioral method to the reward, which means replacing $r_t$ by $r_t + \alpha\ln\pi_0(a_t \mid s_t)$ in any TD scheme. This approach assumes stochastic policy action implemented in entropy-regularized RL approaches. Compared to Munchausen RL, instead of using KL-divergence from the previous policies, we use that from a pretrained Transformer-based behavioral policy to force the RL agent close to the expert demonstrations. In principle, a variety of entropy-regularized RL methods could be extended to SoftCtrl. In this work, we build on SAC known as an actor-critic algorithm using the principle of maximum entropy RL, which performs competitively and stably in continuous tasks.
In SAC-ImKL approach, the critic update in \eqref{eq:sac-critic-target} has been modified as:
\begin{multline}
\label{eq:q-imKL}
    \hat{q}_{\text{SAC-ImKL}}(r_t,s_{t+1}) = r_t+\alpha\tau\ln\pi_0(a_t|s_t)+ \\ \gamma\mathbb{E}_{a^{\prime}\sim{\pi_\theta}}\big[q_{\bar{\phi}}(s_{t+1},a^{\prime})-\tau\ln\pi_{\theta}(a^{\prime}|s_{t+1})\big]
\end{multline}
where $\alpha \in [0,1]$ is a scaling factor. When the limit of $\alpha \rightarrow 0$ we retrieve maximum entropy RL.  The actor update remains the same as in SAC (see in \eqref{eq:sac-actor-update}). What happens underneath this modification is that it implicitly performs KL regularization between the learned policy and a reference policy. This can be accomplished by a reparameterization trick with $q_k^\prime = q_k - \alpha\tau \ln \pi_0 $. With slight abuse of the notion, we write  $\pi_k = \pi(\cdot \mid s_t)$, $q_k = q(r_{t-1},s_t)$, and $\sum\pi q = \mathbb{E}_{a^{\prime}\sim{\pi}}[q]$. In the policy evaluation step, the regression target can be rewritten as:

\begin{equation}
\begin{aligned}
\hat{q}_{k+1}=r+\alpha \tau \ln \pi_0+\gamma \sum\pi_{k+1}\left(q_k-\tau \ln \pi_{k+1}\right) 
\\ \Leftrightarrow \hat{q}_{k+1}-\alpha \tau \ln \pi_0=r+\gamma \sum \pi_{k+1}\bigg(q_k-\alpha \tau \ln \pi_0 
\\ -(1-\alpha) \tau \ln \pi_{k+1}-\alpha \tau \ln \frac{\pi_{k+1}}{\pi_0}\bigg) 
\\ \Leftrightarrow \hat{q}_{k+1}^{\prime}=r+\gamma \sum \pi_{k+1}q_k^{\prime}+(1-\alpha) \tau \mathcal{H}(\pi_{k+1}) 
\\ - \alpha\tau  \mathbb{D}_{\mathrm{KL}}(\pi_{k+1} \| \pi_0)
\end{aligned}
\end{equation}

Then, we also rewrite the policy improvement step $\pi_{k+1} = \arg \max_{\pi\in\Delta_{\cal A}^{\cal S}}\sum\pi q_k + \tau \mathcal{H}(\pi)$  as:
\begin{equation}
\begin{aligned}
    & \sum\pi q_k + \tau \mathcal{H}(\pi)  
    \\& = \sum\pi (q_k^{\prime} + \alpha\tau \ln\pi_0) - \tau \sum\pi\ln\pi 
    \\& = \sum\pi q_k^{\prime} - \alpha\tau \sum\pi(\ln\frac{\pi}{\pi_0}) - (1-\alpha)\tau\sum\pi\ln\pi
    \\& = \sum\pi q_k^{\prime} - \alpha\tau \mathbb{D}_{\mathrm{KL}}(\pi \| \pi_0) + (1-\alpha)\tau\mathcal{H}(\pi)
\end{aligned}
\end{equation}

We have just shown that our algorithm implicitly performs maximum entropy and minimum KL regularization from the reference policy $\pi_0$, with KL scaled by $\alpha\tau$ and entropy scaled by $(1 - \alpha)\tau$.

\subsection{Forward and Inverse Vehicle Dynamics Models}
\label{sec:kin_model}
We update the vehicle's state using the Unicycle Kinematic Model, which computes the next state or pose of the ego SDV $(x, y, \theta)$ given a 2-dimensional input steering and acceleration action $a = (a_{\text{steer}}, a_{\text{accel}})$. We employ an inverse dynamics model that finds expert actions indicating the same ego's pose as the log replay data of ego and agents. This inverse process helps to convert IL's output (states/pose) into 2-dimensional action, allowing IL and RL to share the same action space. Given the last speed $v_{\text{old}}$ and the current pose of the expert, these actions can be derived in a closed-form solution as
\begin{align}
a_{\text{steer}} & = \theta    \\
a_{\text{accel}} & = \eta  \sqrt{x^2 + y^2} - v_{\text{old}},
\end{align}
where, 
\begin{equation*}
\eta= \begin{cases} 1 , & \text { if }x\cos(\theta) > 0 \\ -1, & \text { otherwise }\end{cases}
\end{equation*}
These expert actions are means of Diagonal Gaussian behavioral distribution $\pi_0 = \mathcal{N}(\boldsymbol{\mu}_0, \boldsymbol{\sigma}_0^2)$. We choose constant deviation $\boldsymbol{\sigma}_0 = [\sigma_{\text{steer}}, \sigma_{\text{accel}}] = [\exp(-1.5), \exp(-1.5)] $

\subsection{Reward function}
Similar to \cite{kothari2021drivergym}, we leverage the expert demonstration and also learn to avoid collisions. The reward function comes by combining imitation-based and safety-based rewards
\begin{equation}
r = \sum_i{\lambda_i  r_i},    
\end{equation}
where, $r_i$ indicates $r_{\text{dist}}$, $r_{\text{yaw}}$, $r_{\text{cf}}$, $r_{\text{cs}}$, and $r_{\text{cr}}$, which are describe as follows:

\begin{align}
r_{\text{dist}} &= -\max \left(\left \|\mathbf{p}_{\text {pred }}-\mathbf{p}_{\text{gt}}\right\|_2 , 20 \right )\\
r_{\text{yaw}} &=-\|\theta_{\text{pred}} - \theta_{\text{gt}}\|_2 \\
r_{\text{cf}} & =r_{\text{cs}}=r_{\text{cr}}=-1,
\end{align}
where $\mathbf{p}_{\text{pred}}$, $\mathbf{p}_{\text{gt}}$ respectively indicate the predicted position $(x,y)$ of the RL agent and ground-truth ego replay at every time step. For the distance-based rewards, we use L2-norm. Reward clipping is performed in $r_{\text{dist}}$ with a maximum threshold of 20 for stability in RL optimization. The heading angle $\theta$ is bounded, thus $r_{\theta}$ does not need to perform reward clipping. $r_{\text{cf}}, r_{\text{cs}}, r_{\text{cr}}$ are the rewards of collision front, collision side, and collision rear respectively at every time step. All of safety-based rewards are set to $-1$ to punish the agent and their weights are set to equal to the maximum value (20) of imitation cost. The weight $\lambda$ corresponding to each reward is shown in Table~\ref{tab:reward_weight}.

\begin{table}[htbp]
\caption{Reward weights.}
\centering
\begin{tabular}{cc}
\hline
\textbf{Reward} & \textbf{Weight $\lambda$} \\ \hline
distance            & 1.0             \\ \hline
yaw             & 1.0             \\ \hline
collision front & 20.0             \\ \hline
collision side  & 20.0             \\ \hline
collision rear  & 20.0             \\ \hline
\end{tabular}
\label{tab:reward_weight}
\end{table}
\section{Experiments}

\subsection{Experimental Setup}
\subsubsection{Datasets} We use a dataset consisting of around 100 hours of expert driving trajectories, split into 25-second segments. All logs were collected by a fleet of 20 self-driving vehicles driving along a fixed route in Palo Alto, California \cite{houston2021one}. The training data comprises 16K scenes where the initial frame is randomly sampled. We test the performance of models on 100 unseen segments. The state representation captures a collection of static and dynamic elements around the ego. In particular, static elements include traffic lanes, stop signs, and pedestrian crossings, which are retrieved from the localization system. The dynamic elements including traffic light status and surrounding agents (other cars, buses, pedestrians, and cyclists) are detected in real time using the onboard perception system. Regarding to ego state, we utilize vectorized form derived directly from HD maps elements which contain agent dynamics and structured scene context. Each feature of a vector map can be represented as multiple points, polygons, or curves. The input and output of the model we used are similar to \cite{scheel2022urban}, which utilizes past and current states of all elements including ego car to predict the future states of the ego. Therefore, no navigation information that guides the vehicle toward the destination is given.

\subsubsection{Simulation}
The simulator is based on the OpenAI Gym framework and uses a physics-based simulation engine to model the behavior of an ego vehicle in various driving scenarios. As mentioned in section~\ref{sec:kin_model}, the simulator uses the Unicycle Kinematic Model to simulate the SDV ego's motion. On the other hand, the trajectories of other agents and pedestrians in the scene are replayed from the logs (log-playback), similarly to \cite{kothari2021drivergym, lu2023imitation}. Although this implies that the agents are non-reactive, it guarantees that the actions of other agents resemble human behavior. The simulation also allow new agents to enter ego-centered frames, resulting in more complicated scenarios. The average number of agents per frame in the train and test set is $79$ and $76$, respectively.

\begin{table*}[htbp]
\centering
\caption{Evaluation of different methods using closed-loop evaluation in DriverGym on 100 sampled scenarios. Lower is better. The values in the table are expressed as mean $\pm$ standard deviation. Our SAC-ImKL methods improve safety over BC-perturb baselines by reducing the collision rate while performing as smoothly as BC-perturb.}
\label{tab:quantitative}
\begin{tabular}{|c|ccc|cc|cc|c|}
\hline
\multirow{2}{*}{\textbf{Method}} & \multicolumn{3}{c|}{\textbf{Imitation}}                                                                                                                    & \multicolumn{2}{c|}{\textbf{Collisions}}                                                         &  \multicolumn{2}{c|}{\textbf{Discomfort}}& \textbf{Failure}                           \\ \cline{2-9} 
                                 & \multicolumn{1}{c|}{$\mu_{\text{ADE}}$}                     & \multicolumn{1}{c|}{$\mu_{\text{D2R}}$}              & D2R ($\ge4$m)                         & \multicolumn{1}{c|}{$\mu_{\text{CL}}$}             & CL ($\ge 1$)                        &  \multicolumn{1}{c|}{$\mu_{\text{ACC}}$}&ACC ($\ge2$m/s\textsuperscript{2}) & D2R+CL                                 \\ \hline
BC-perturb                              & \multicolumn{1}{c|}{$\mathbf{10.63 \pm \scriptstyle 0.06}$} & \multicolumn{1}{c|}{$\mathbf{1 \pm \scriptstyle 0}$} & $8.67 \pm \scriptstyle 0.58$ & \multicolumn{1}{c|}{$8 \pm \scriptstyle 0.1$}              & $25.33 \pm \scriptstyle 1.53$       &  \multicolumn{1}{c|}{$0.05 \pm \scriptstyle 0.00$} &$3.86 \pm \scriptstyle 0.26$                            & $34 \pm \scriptstyle 2$                \\
BPTT& \multicolumn{1}{c|}{$19.9 \pm \scriptstyle 1.2$} & \multicolumn{1}{c|}{$1.4 \pm \scriptstyle 0.3$} & $\mathbf{6 \pm \scriptstyle 0.5}$& \multicolumn{1}{c|}{$10 \pm \scriptstyle 0.82$}              & $49 \pm \scriptstyle 3.43$&  \multicolumn{1}{c|}{$0.04 \pm \scriptstyle 0.01$} &$3.67 \pm \scriptstyle 0.12$& $55 \pm \scriptstyle 4$\\
\hline
PPO                              & \multicolumn{1}{c|}{$45.5 \pm \scriptstyle 12.91$}           & \multicolumn{1}{c|}{$1.63 \pm \scriptstyle 0.4$}    & $15.33 \pm \scriptstyle 5.13$         & \multicolumn{1}{c|}{$20.87 \pm \scriptstyle 1.14$}          & $139 \pm \scriptstyle 27.5$          &  \multicolumn{1}{c|}{$0.03 \pm \scriptstyle 0.01$} &$1.06 \pm \scriptstyle 0.43$& $154.33 \pm \scriptstyle 22.9$          \\
SAC                              & \multicolumn{1}{c|}{$24.3 \pm \scriptstyle 2.05$}           & \multicolumn{1}{c|}{$2.43 \pm \scriptstyle 0.15$}    & $21.33 \pm \scriptstyle 2.08$         & \multicolumn{1}{c|}{$1.87 \pm \scriptstyle 0.25$}          & $27 \pm \scriptstyle 5.29$          &  \multicolumn{1}{c|}{$0.09 \pm \scriptstyle 0.04$} &$13.61 \pm \scriptstyle 10.64$                          & $48.33 \pm \scriptstyle 6.66$          \\
SAC-ExKL                         & \multicolumn{1}{c|}{$21.03 \pm \scriptstyle 1.3$}           & \multicolumn{1}{c|}{$2.43 \pm \scriptstyle 0.15$}    & $15.33 \pm \scriptstyle 2.31$         & \multicolumn{1}{c|}{$1.03 \pm \scriptstyle 0.61$}          & $19.33 \pm \scriptstyle 7.57$       &  \multicolumn{1}{c|}{$\mathbf{0.02 \pm \scriptstyle 0.00}$} &\textbf{$\mathbf{0.31 \pm \scriptstyle 0.06}$}                   & $34.67 \pm \scriptstyle 6.43$              \\
SAC-ImKL                         & \multicolumn{1}{c|}{$22.57 \pm \scriptstyle 0.71$}          & \multicolumn{1}{c|}{$2.1 \pm \scriptstyle 0.1$}      & $13.33 \pm \scriptstyle 2.52$         & \multicolumn{1}{c|}{$\mathbf{0.83 \pm \scriptstyle 0.06}$} & $\mathbf{15 \pm \scriptstyle 2.65}$ &  \multicolumn{1}{c|}{$0.05 \pm \scriptstyle 0.01$}&$3.83 \pm \scriptstyle 1.89$                            & $\mathbf{28.33 \pm \scriptstyle 0.58}$ \\ \hline
\end{tabular}

\label{tab:quantitative}
\end{table*}
\subsubsection{Model architecture}
For our urban driving scenarios, we discovered that SAC-ImKL with 2 separate feature extractor networks between actor and critic net yielded superior overall results. This separation allows each network to focus on different aspects of the input data, thus leading to more accurate representations for each network. To boost RL training, we initialized backbones weights with pretrained perturbation BC model \cite{bansal2018chauffeurnet}. 
\subsection{Baselines}
We compare our proposed algorithms against these approaches.
\begin{itemize}
    \item \textbf{BC-perturb}: We re-implement BC with perturbations \cite{bansal2018chauffeurnet} using Transformer backbone of \cite{scheel2022urban} to represent open loop IL method. This model directly predicts the most likelihood of the next $T=12$ ego future poses $[(x_{1}, y_{1},\theta_{1}), \ldots (x_{T}, y_{T},\theta_{T})]$. 
    \item \textbf{BPTT}: We re-implement IL with backpropagation through time \cite{scheel2022urban} using Transformer backbone of \cite{scheel2022urban} to represent closed-loop IL method. Similar to BC-perturb. this model predicts the next $T=12$ ego future poses.
    \item \textbf{PPO}: We also develop a Proximal Policy Optimization (PPO) baseline to represent on-policy RL-only approaches that perform optimally with frozen and shared Transformer backbone weights initialized by a pretrained BC model with perturbations.
    \item \textbf{SAC}: We also include a SAC baseline as an off-policy RL-only approach. The backbone of this approach is the same as that of SAC-ImKL.
    \item \textbf{SAC-ExKL}: We also implement a direct KL control for the auto-tuned entropy version of SAC. This approach explicitly adds KL to the SAC's reward function. The backbone of this approach is the same as that of SAC-ImKL.
\end{itemize}
All IL and RL methods share the same observation input described in \cite{scheel2022urban}. We fine-tune PPO, SAC, SAC-ExKL, and SAC-ImKL on 6M, 300K, 200K, and 200K steps respectively. Based on the results, we picked the checkpoints with the highest scores on the validation set, as our final models.
\subsection{Metrics}
In this work, we use a large-scale real-world dataset with closed-loop simulation to evaluate our SDV planning model performance. For each segment in the test set, we simulated the SDV by its learned policy for the full duration of a segment and detected continuous or binary events including expert imitation, safety, and comfort of ego car, which are detailed as follows:
\begin{itemize}
    \item \textbf{Average Displacement Error (ADE):} Computes the L2 distance between predicted ego centroid and ground-truth ego centroid averaged over the entire episode. 
    \item \textbf{Distance To Reference (D2R):} Computes the L2 distance between the predicted centroid and the closest waypoint in the reference trajectory (ground-truth ego). We raise an off-road failure event if the simulated SDV deviates from the reference trajectory center line by greater than $4$m.
    \item \textbf{Collision (CL):}  Collisions occurred between the ego and any other agent, divided into collision front (CF), side (CS), and rear (CR). For validation, we raise a collision failure event for a scene if have at least one collision in that scene segment.
    \item \textbf{Discomfort (ACC):} Percentage in a frame of recording the absolute value of acceleration raises a failure should this exceed $2$m/s\textsuperscript{2}.
    \item \textbf{Failure (D2R+CL):} We combine collisions and off-road events into one metric for ease of comparison.
\end{itemize}
Note that all metrics are computed per timestep, normalizing by the number of scenes in the validation test set. The normalized average values of ADE, D2R, ACC, CL, CF, CS, and CR are denoted as $\mu_{\text{ADE}}$, $\mu_{\text{D2R}}$, $\mu_{\text{ACC}}$, $\mu_{\text{CL}}$, $\mu_{\text{CF}}$, $\mu_{\text{CS}}$ and $\mu_{\text{CR}}$ respectively. Compared to \cite{scheel2022urban} process of resetting after interventions, and the use of a segment of 10 seconds in \cite{lu2023imitation}, we use a much harder validation test that in a whole 25 seconds segment, we cumulatively count for any instance of off-road or collision event in binary format results in a failure including D2R ($\ge4$m), ACC  ($\ge2$m/s\textsuperscript{2}), CL ($\ge 1$), CF ($\ge 1$), CS ($\ge 1$), and CR ($\ge 1$). Therefore, the ego is allowed to drive to the end-of-segment frame without any intervention of failure, which potentially causes more failures.


\subsection{Results}
Following the above derivations, we first conduct a comprehensive comparison between our SAC-ImKL and 4 baselines (BC-perturb, BPTT, PPO, SAC, SAC-ExKL). Table~\ref{tab:quantitative} reports the performance of the urban driving rules to which the models are trained to adhere, for example,  keeping progress (ADE), avoiding collisions and off-route events (Failure), and driving smoothly like humans (Discomfort). Each configuration uses a mean and standard deviation derived from 3 random seeds. Many hyperparameters could be tuned to further improve the metrics; however, we found that $\sigma_{\text{steer}} =  \sigma_{\text{accel}} = \exp(-1.5)$, $\alpha = 0.3$ for SAC-ExKL,  $\tau = 1.2$, $\alpha = 0.4$ for SAC-ImKL produce satisfied planners. 

\subsubsection{Comparing regularized RL with IL-only and RL-only}
\begin{figure}[!t]
    \begin{center}
    \includegraphics[width=\linewidth]{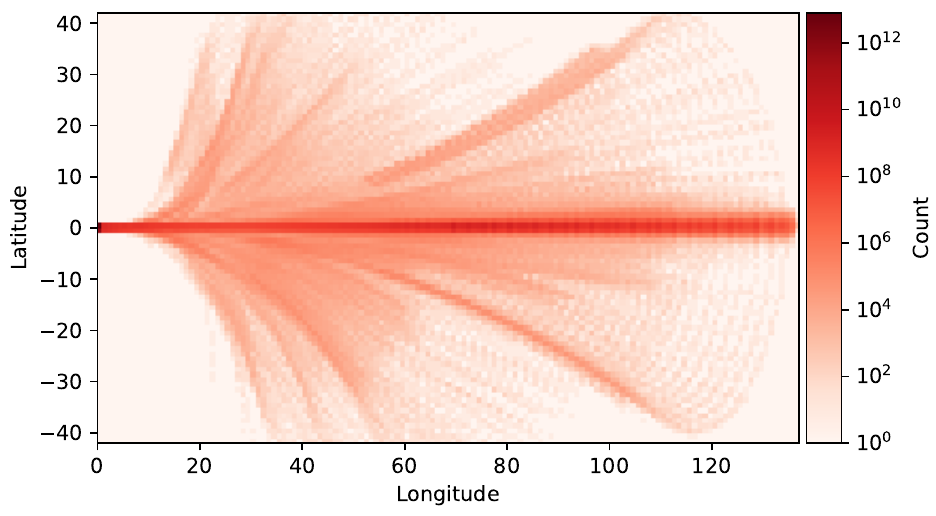}
    \end{center}
    \caption[trajectory heatmap]{Distribution of ego vehicle’s future location in the Lyft Dataset \cite{houston2021one}.}
    \label{fig:trajectory_heatmap}
\end{figure}
In Table~\ref{tab:quantitative}, BC-perturb and BPTT outperform RL in imitation tasks since 1) IL directly mimics human driving behavior by supervised learning, which is easier than RL letting the agent explore and gives a reward whenever the agent drives properly and 2) in multiple possible outcomes scenarios like the T-junction or crossroad, RL can deviate from logs since it allows the agent to explore more options. However, the two IL methods yield higher average collision $\mu_{CL}$ than off-policy RLs likely due to covariate shift \cite{ross2011reduction}. We also hypothesize that RL is more robust to highly biased human data than IL. As evidence, we visualize the attention heatmap (Fig.~\ref{fig:attention_map}) of all approaches over 4 different types of elements: ego's past trajectories, agent past and current trajectories, lane, and crosswalk in 30 sampled frames. The figure demonstrates that BC-perturb almost exclusively attends to its past trajectories, similar to human distribution that bias on simple cases in which future trajectories could be extrapolated by their past trajectories shown in Fig.~\ref{fig:trajectory_heatmap}. In contrast, RL methods tend to utilize more information from surrounding traffic elements. A more detailed discussion is provided in Section~\ref{sec:rl-help-transformer}).

\begin{figure}[htbp]
    \centering
    \begin{subfigure}{.23\textwidth}
         \centering
         \includegraphics[width=\textwidth]{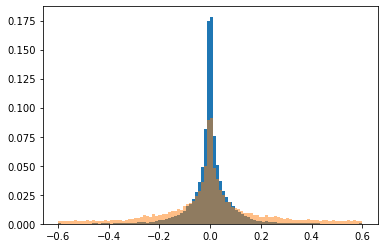}
         \caption{SAC acceleration}
    \end{subfigure}
    \begin{subfigure}{.23\textwidth}
        \centering
        \includegraphics[width=\textwidth]{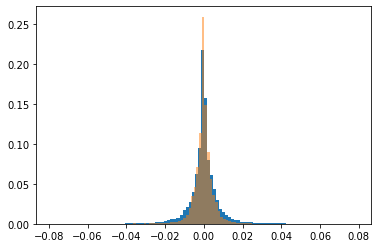}
        \caption{SAC steering angle}
    \end{subfigure}
    \begin{subfigure}{.23\textwidth}
        \centering
        \includegraphics[width=\textwidth]{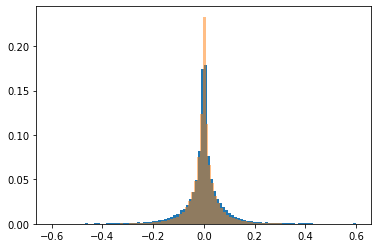}
        \caption{SAC-ExKL acceleration}
    \end{subfigure}
    \begin{subfigure}{.23\textwidth}
        \centering
        \includegraphics[width=\textwidth]{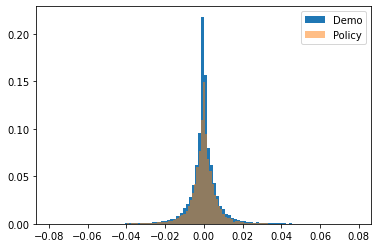}
        \caption{SAC-ExKL steering angle}
    \end{subfigure}
    \begin{subfigure}{.23\textwidth}
        \centering
        \includegraphics[width=\textwidth]{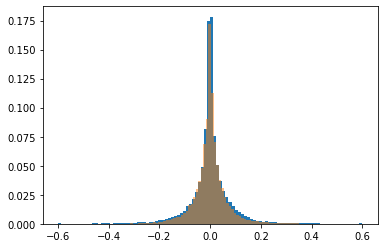}
        \caption{SAC-ImKL acceleration}
    \end{subfigure}
    \begin{subfigure}{.23\textwidth}
        \centering
        \includegraphics[width=\textwidth]{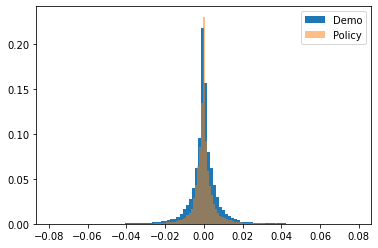}
        \caption{SAC-ImKL steering angle}
    \end{subfigure}
    \caption{Marginal action distributions of SAC/SAC-ExKL/SAC-ImKL (orange) vs replay logs (blue)}
    \label{fig:action_dist}
\end{figure}

\begin{figure}[htbp]
    \begin{center}
    \includegraphics[width=\linewidth]{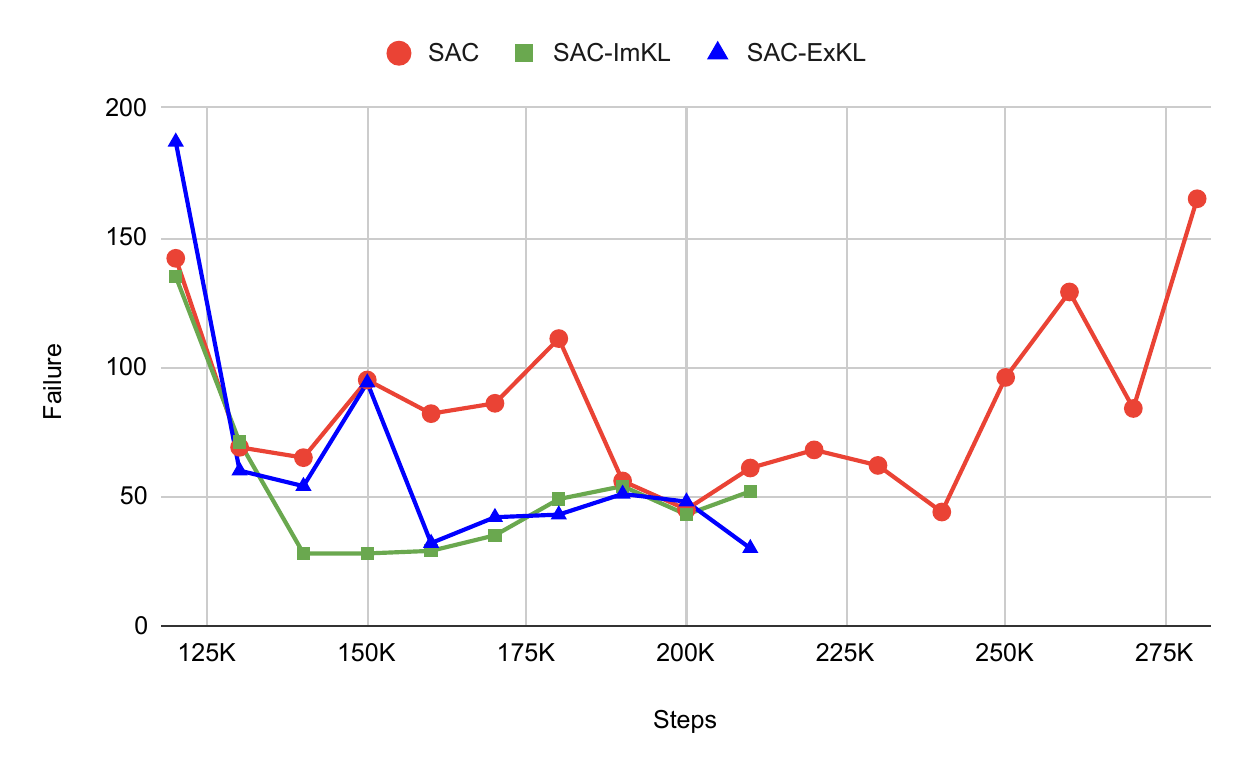}
    \end{center}
        \caption[Failure as a function of timesteps.]{Failure as a function of timesteps of RL methods.}
    \label{fig:failure_vs_timesteps}
\end{figure}

\begin{figure*}[htbp]
     \centering
     \begin{subfigure}{\textwidth}
     \centering
     \begin{minipage}{0.05\textwidth}
        \centering
        \rotatebox{90}{\textbf{BC-perturb}}
    \end{minipage}
     \begin{minipage}{.70\textwidth}
         \centering
         \includegraphics[width=\textwidth]{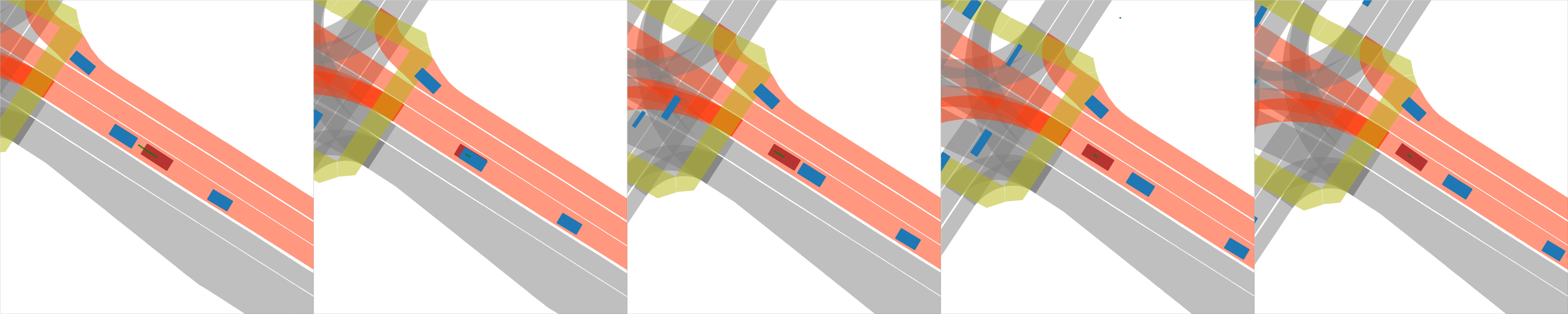}
     \end{minipage}
     \hfill
    \begin{minipage}{0.01\textwidth}
        \centering
        \hspace*{1mm}
        \rotatebox{90}{\texttt{\tiny{accel}}}
    \end{minipage}
    \begin{minipage}{.19\textwidth}
         \centering
         \includegraphics[width=\textwidth]{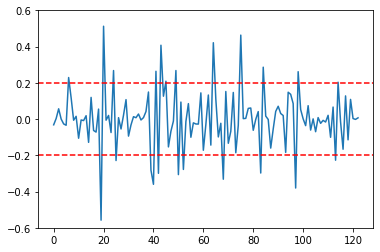}
    \end{minipage}
     \hfill
     \begin{minipage}{0.05\textwidth}
        \centering
        \rotatebox{90}{\textbf{SAC}}
    \end{minipage}
    \begin{minipage}{.70\textwidth}
         \centering
         \includegraphics[width=\textwidth]{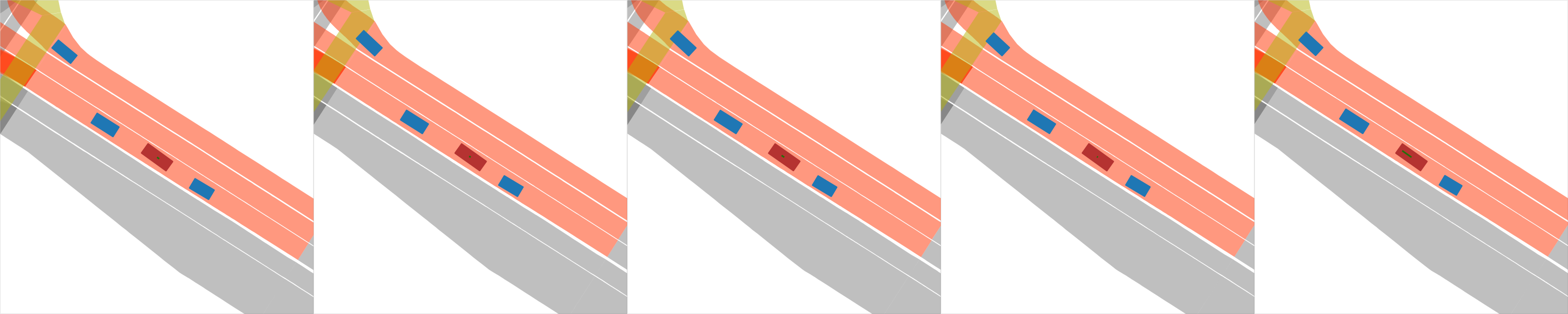}
     \end{minipage}
     \hfill
    \begin{minipage}{0.01\textwidth}
        \centering
        \hspace*{1mm}
        \rotatebox{90}{\texttt{\tiny{accel}}}
    \end{minipage}
     \begin{minipage}{.19\textwidth}
         \centering
         \includegraphics[width=\textwidth]{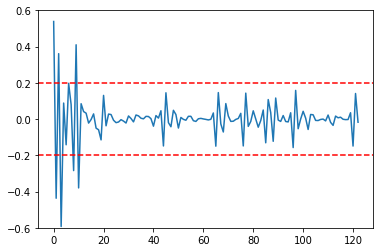}
     \end{minipage}
     \hfill
     \begin{minipage}{0.05\textwidth}
        \centering
        \rotatebox{90}{\textbf{SAC-ExKL}}
    \end{minipage}
     \begin{minipage}{.70\textwidth}
         \centering
         \includegraphics[width=\textwidth]{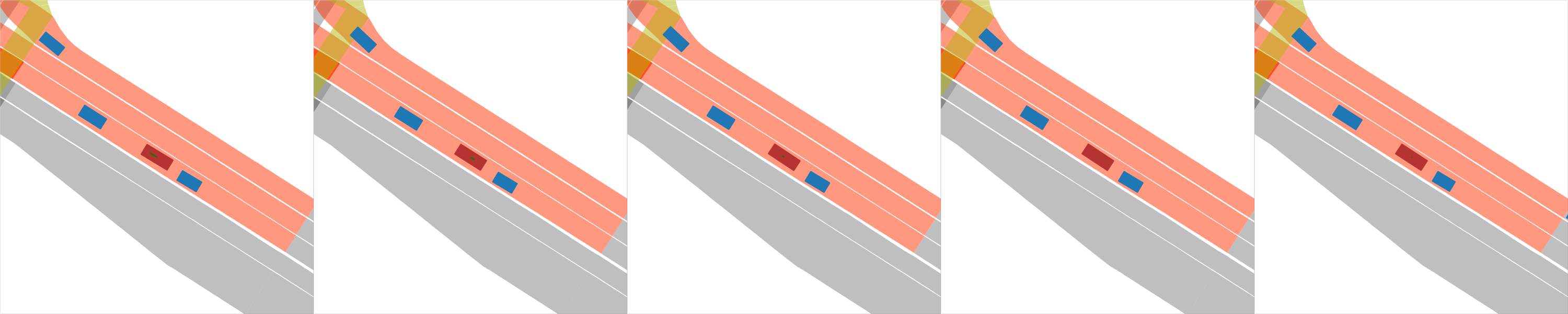}
     \end{minipage}
     \hfill
     \begin{minipage}{0.01\textwidth}
        \centering
        \hspace*{1mm}
        \rotatebox{90}{\texttt{\tiny{accel}}}
    \end{minipage}
     \begin{minipage}{.19\textwidth}
         \centering
         \includegraphics[width=\textwidth]{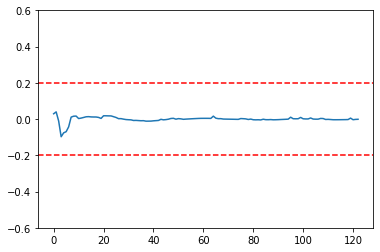}
     \end{minipage}
     \hfill
     \begin{minipage}{0.05\textwidth}
        \centering
        \rotatebox{90}{\textbf{SAC-ImKL}}
    \end{minipage}
     \begin{minipage}{.70\textwidth}
         \centering
         \includegraphics[width=\textwidth]{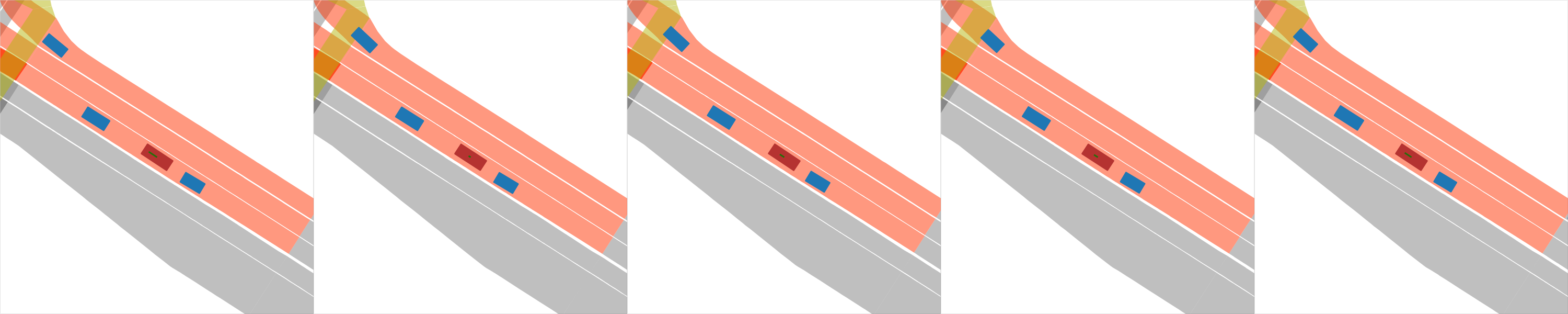}
     \end{minipage}
     \hfill
     \begin{minipage}{0.01\textwidth}
        \centering
        \hspace*{1mm}
        \rotatebox{90}{
        \texttt{\tiny{accel}}}
    \end{minipage}
     \begin{minipage}{.19\textwidth}
         \centering
         \includegraphics[width=\textwidth]{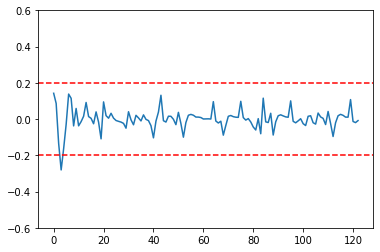}
         \vspace*{-8mm} 
         \caption*{\texttt{\tiny{frame}}}
     \end{minipage}
               \caption{Red traffic light scene}
     \end{subfigure}

    \begin{subfigure}{\textwidth}
    \centering
    \begin{minipage}{0.05\textwidth}
        \centering
        \rotatebox{90}{\textbf{BC-perturb}}
    \end{minipage}
     \begin{minipage}{.70\textwidth}
         \centering
         \includegraphics[width=\textwidth]{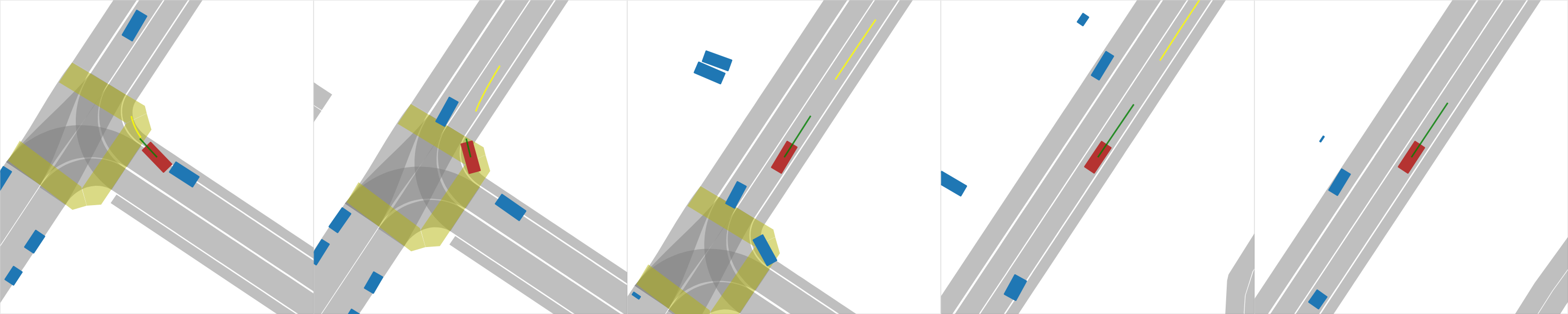}
     \end{minipage}
     \hfill
     \begin{minipage}{0.01\textwidth}
        \centering
        \hspace*{1mm}
        \rotatebox{90}{
        \texttt{\tiny{accel}}}
    \end{minipage}
    \begin{minipage}{.19\textwidth}
         \centering
         \includegraphics[width=\textwidth]{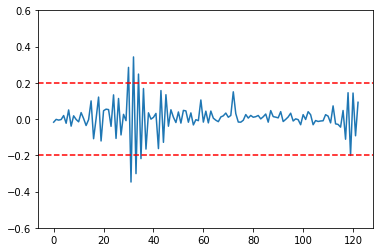}
    \end{minipage}
     \hfill
     \begin{minipage}{0.05\textwidth}
        \centering
        \rotatebox{90}{\textbf{SAC}}
    \end{minipage}
    \begin{minipage}{.70\textwidth}
         \centering
         \includegraphics[width=\textwidth]{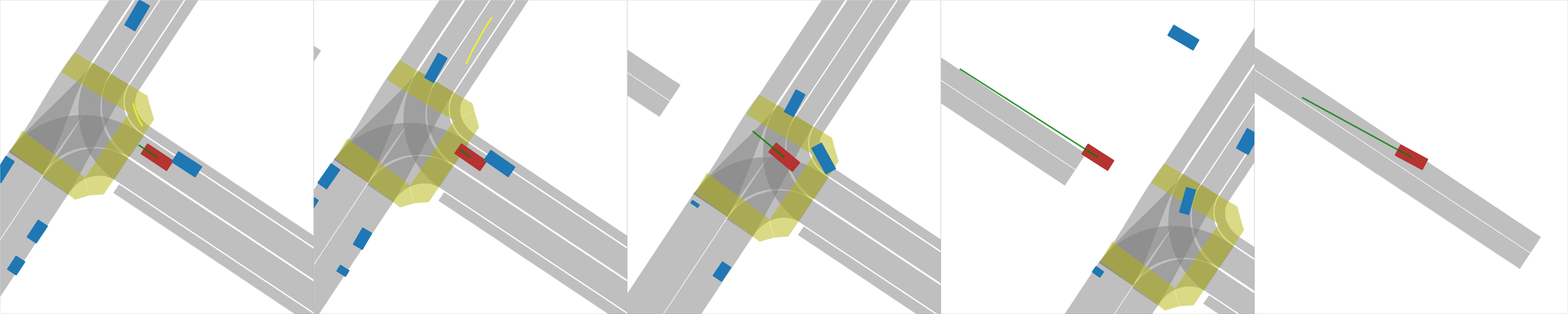}
     \end{minipage}
     \hfill
     \begin{minipage}{0.01\textwidth}
        \centering
        \hspace*{1mm}
        \rotatebox{90}{
        \texttt{\tiny{accel}}}
    \end{minipage}
     \begin{minipage}{.19\textwidth}
         \centering
         \includegraphics[width=\textwidth]{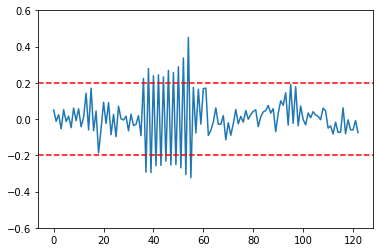}
     \end{minipage}
     \hfill
     \begin{minipage}{0.05\textwidth}
        \centering
        \rotatebox{90}{\textbf{SAC-ExKL}}
    \end{minipage}
     \begin{minipage}{.70\textwidth}
         \centering
         \includegraphics[width=\textwidth]{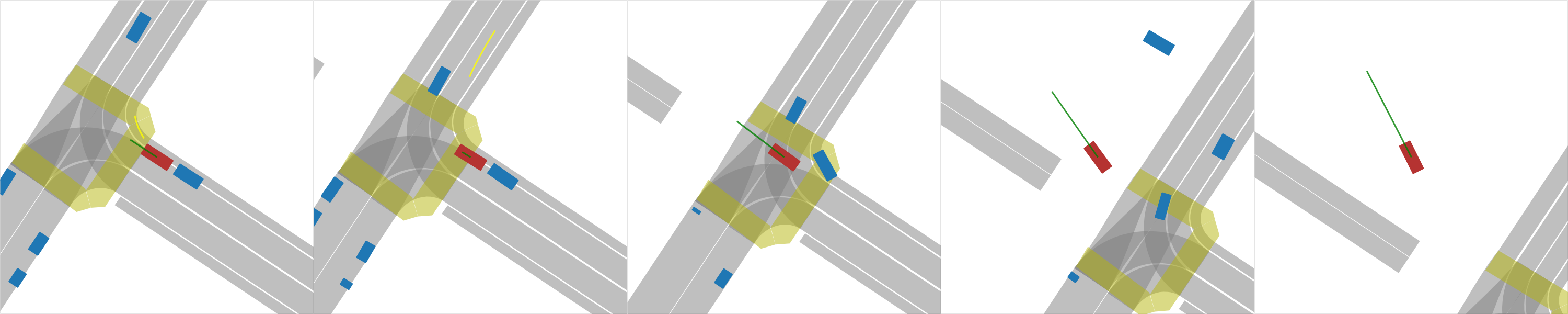}
     \end{minipage}
     \hfill
     \begin{minipage}{0.01\textwidth}
        \centering
        \hspace*{1mm}
        \rotatebox{90}{
        \texttt{\tiny{accel}}}
    \end{minipage}
     \begin{minipage}{.19\textwidth}
         \centering
         \includegraphics[width=\textwidth]{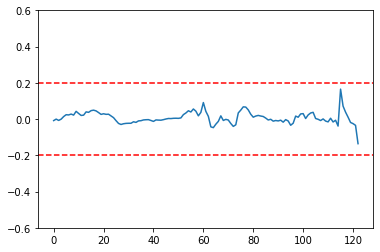}
     \end{minipage}
     \hfill
     \begin{minipage}{0.05\textwidth}
        \centering
        \rotatebox{90}{\textbf{SAC-ImKL}}
    \end{minipage}
     \begin{minipage}{.70\textwidth}
         \centering
         \includegraphics[width=\textwidth]{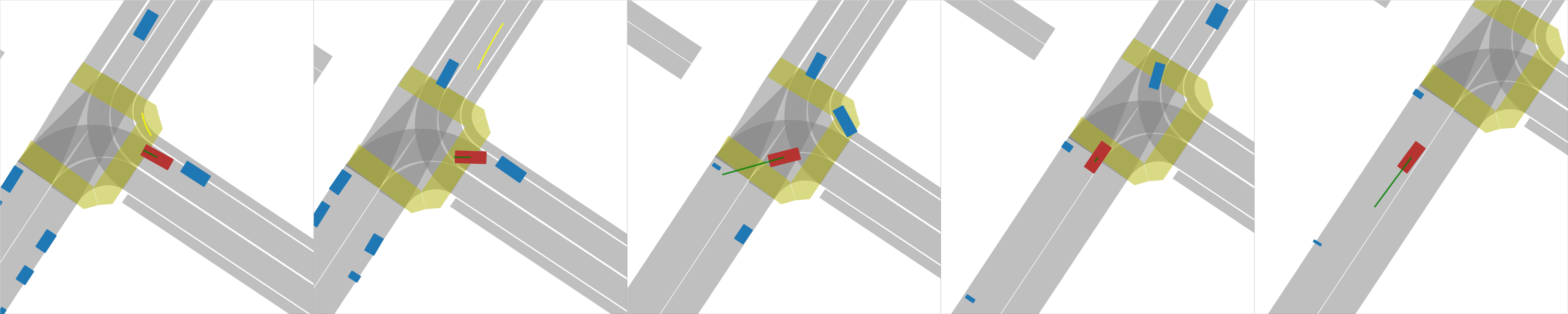}
     \end{minipage}
     \hfill
     \begin{minipage}{0.01\textwidth}
        \centering
        \hspace*{1mm}
        \rotatebox{90}{
        \texttt{\tiny{accel}}}
    \end{minipage}
     \begin{minipage}{.19\textwidth}
         \centering
         \includegraphics[width=\textwidth]{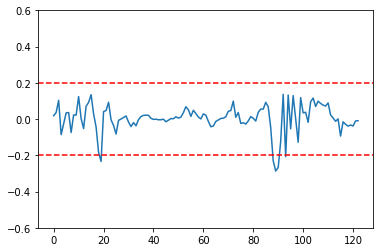}
         \vspace*{-8mm} 
         \caption*{\texttt{\tiny{frame}}}
     \end{minipage}
     \caption{T-junction scene}
     \end{subfigure}
    \caption[Demostrations of 2 win case against baselines method in the unseen test set.]{Demostrations of 2 win case against baselines method in the unseen test set. Every row depicts 5 frames in 1 scene, and images are 2.5s apart (SDV in red, other agents in blue, crosswalks in yellow, SDV trajectory in the green line, SDV ground-truth trajectory in yellow line). The right side plots are the associated acceleration action $a_{\text{accel}}$  (red dashed lines represent the boundary value of smoothness). \textbf{Scenario 1}: BC-perturb collides with a front vehicle stopped at a red light, while RL methods stop and leave enough clearance. \textbf{Scenario 2:} RL diverges from logs in the T-junction case. Despite SAC-ImKL results in a higher IL error than BC-perturb, it executes a proper and smooth left turn without a collision.}
    \label{fig:quanlitative}
    \end{figure*}

In our evaluation setting, BC-perturb outperforms BPTT overall; therefore, we prefer to use BC-perturb as IL-alone to compare with other approaches. PPO achieves better $\mu_{D2R}$, D2R than SAC variants, however, it is the worst candidate overall due to the high collision rate, which shows the drawback of RL policy in a complex urban driving task. SAC, an off-policy RL, outperforms BC-perturb in average collision per scene, however, it generates more collision cases (CL) and discomfort trajectories. This is because the entropy term encourages the policy to be uniform, yielding unnatural (e.g., swerves) and uncomfortable (e.g., abrupt acceleration/deceleration) driving behavior, as shown in Fig.~\ref{fig:action_dist}. With KL regularizer, SAC-ImKL and SAC-ExKL generate policy distributions similar to log trajectories, significantly reducing D2R, CL, and ACC due to becoming more human while improving the sample efficiency of RL in challenging environments (Fig.~\ref{fig:failure_vs_timesteps}).
Fig.~\ref{fig:quanlitative} shows two scenarios in which SAC-ImKL overall wins against the baseline agents.

\begin{figure}[htbp]
    \begin{center}
    \includegraphics[width=\linewidth]{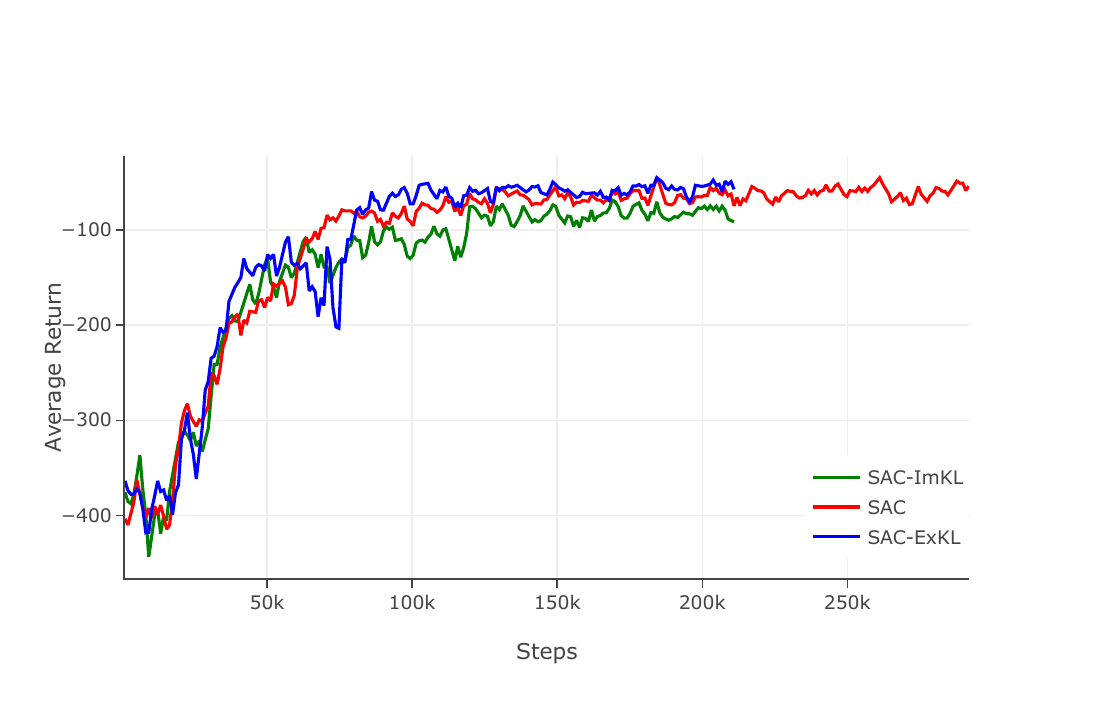}
    \end{center}
        \caption[Training curves on urban navigation DriverGym benchmarks.]{Training curves on urban navigation DriverGym benchmarks.}
    \label{fig:reward-curve}
\end{figure}

\subsubsection{Effect of entropy term in reducing over-conservation in Driving}

Entropy in SAC-ExKL has an uncontrollable influence compared to SAC-ImKL since the former approach uses auto-tuned entropy, which discourages the ability to search for the best trade-off between entropy and KL. In contrast, the latter method provides a generalized entropy-KL RL extended from SAC, which is simple to tune entropy and KL. In Table~\ref{tab:quantitative}, our SAC-ImKL achieves the best performance, reducing $17\%$ failures compared to BC and SAC-ExKL baselines while driving as smoothly as BC policy. 
Fig.~\ref{fig:reward-curve} shows the average returns of the evaluation rolls during training for SAC, SAC-ExKL, and SAC-ImKL, each performing an evaluation rollout every $10$ environment step. In general, SAC-ImKL has a worse asymptotic reward curve compared to the others. Nevertheless, the validation score in Fig.~\ref{fig:failure_vs_timesteps} indicates the opposite trend that SAC-ImKL with tunable entropy coefficient achieves the fewest failures in the shortest time compared to SAC and SAC-ExKL. This contradictory phenomenon is anticipated given that human-designed rewards are inherently imperfect for evaluation. Furthermore, entropy plays a critical role in improving performance and comfort. The sharpening or softening level of the action policy is obtained using a low or a high value for the entropy temperature $\tau$ in the softmax of the Q function in \eqref{eq:pi_soft_objective}. As experimentally shown in Fig.~\ref{fig:entropy-effect-of-SAC-ImKL}, a softer policy gives better performance than a sharper one. The reason is that low entropy and high KL divergence encourage dominance in a solution, while high entropy prevents RL policy from collapsing into a solution and biases the policy toward a uniform distribution. Therefore, increasing diversity via entropy regularization is essential for escaping from the suboptimal solution of its BC teacher.



 \begin{table*}[htbp]
\centering
\caption{In-depth collisions and smoothness on 100 sampled scenarios. Lower is better. SAC-ImKL generates fewer front and side collisions but has a slightly greater frequency of being hit by other objects than SAC-ExKL. }
\label{tab:in-depth-collision}
\begin{tabular}{|c|c|c|c|c|c|c|}
\hline
\multicolumn{1}{|c|}{Method} & $\mu_{\text{CF}}$            & $\mu_{\text{CR}}$            & $\mu_{\text{CS}}$            & CF ($\ge 1$)                 & CR ($\ge 1$)                  & CS ($\ge 1$)                  \\ \hline
BC-perturb& $2.47 \pm \scriptstyle 0.31$ & $2.97 \pm \scriptstyle 0.12$ & $2.57 \pm \scriptstyle 0.15$ & $4.67 \pm \scriptstyle 0.58$ & $10.67 \pm \scriptstyle 0.58$ & $10.00 \pm \scriptstyle 1.00$ \\
SAC                    & $0.27 \pm \scriptstyle 0.21$ & $0.17 \pm \scriptstyle 0.21$ & $1.43 \pm \scriptstyle 0.51$ & $5.67 \pm \scriptstyle 2.52$ & $6.33 \pm \scriptstyle 2.08$  & $15.00 \pm \scriptstyle 2.65$ \\
SAC-ExKL               & $0.13 \pm \scriptstyle 0.06$ & $\mathbf{0.03 \pm \scriptstyle 0.06}$& $0.87 \pm \scriptstyle 0.50$ & $4.33 \pm \scriptstyle 2.31$ & $\mathbf{3.00 \pm \scriptstyle 2.65}$& $12.00 \pm \scriptstyle 2.65$ \\
SAC-ImKL               & $\mathbf{0.00 \pm \scriptstyle 0.00}$& $0.17 \pm \scriptstyle 0.12$ & $\mathbf{0.67 \pm \scriptstyle 0.15}$& $\mathbf{2.00 \pm \scriptstyle 0.00}$& $3.67 \pm \scriptstyle 1.53$  & $\mathbf{9.33 \pm \scriptstyle 1.15}$\\ \hline
\end{tabular}
\end{table*}



\subsubsection{The trade-off between the KL and the entropy terms}
\label{sec:kl-entropy-tradeoff}
\begin{figure}[htbp]
\centering
\begin{subfigure}{.48\textwidth}
    \begin{center}
    \includegraphics[width=\linewidth]{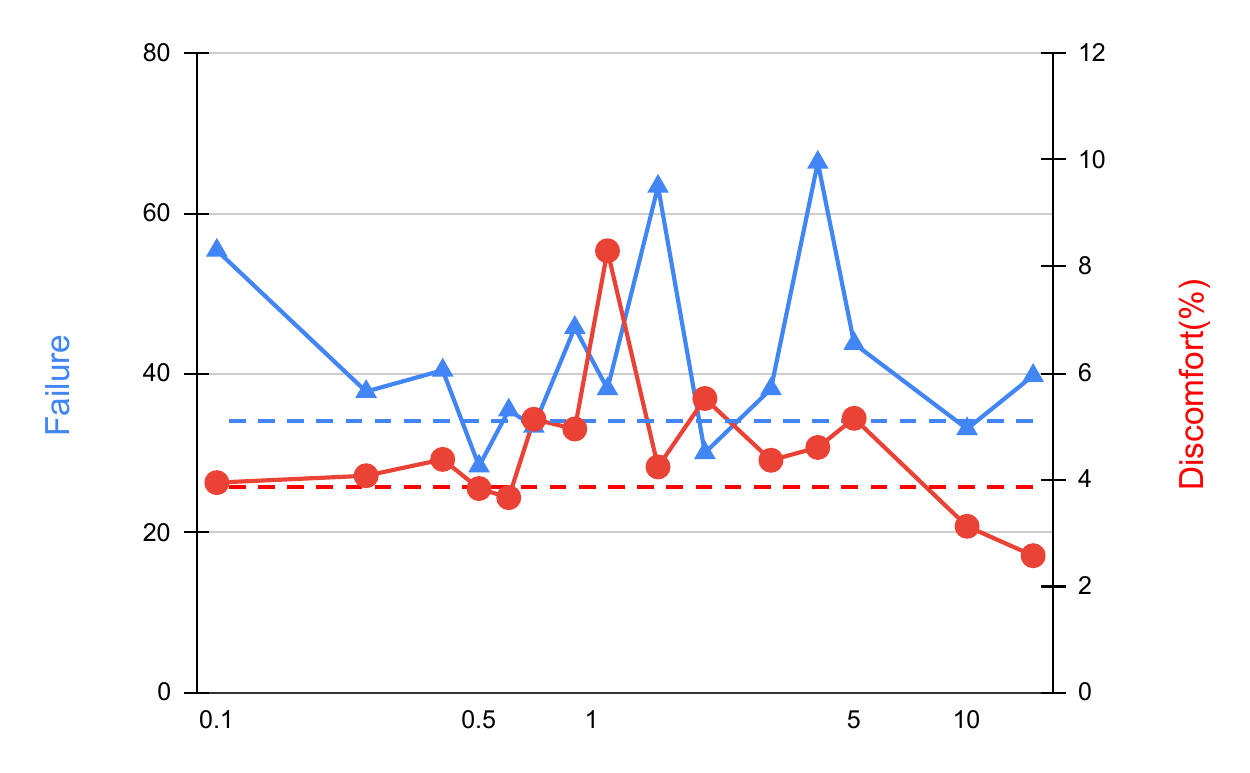}
    \end{center}
        \caption{KL coefficient $w_{\text{KL}}$ ($w_{\cal H}=0.7$)}
    \label{fig:KL-effect-of-SAC-ImKL}

\end{subfigure}
\begin{subfigure}{.48\textwidth}
    \begin{center}
    \includegraphics[width=\linewidth]{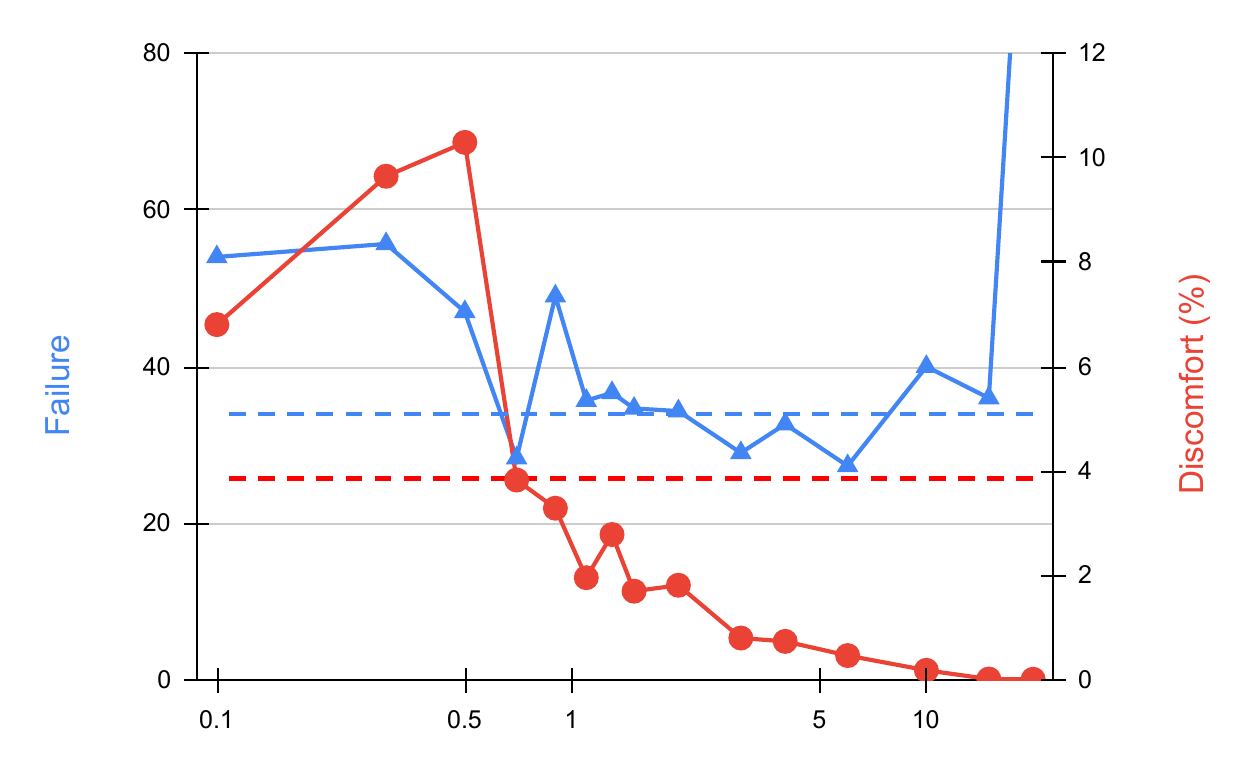}
    \end{center}
        \caption{Entropy coefficient $w_{\cal H}$ ($w_{\text{KL}}=0.5$)}
    \label{fig:entropy-effect-of-SAC-ImKL}
\end{subfigure}
\caption{The effect of KL and entropy terms of SAC-ImKL that are evaluated on the validation set (Failure and Discomfort). The blue and red dash lines indicate the Failure and Discomfort of BC-perturb respectively.}
\end{figure}

This section will focus on SAC-ImKL since it allows us to control the affection of both entropy and KL terms while SAC-ExKL does not. We implement experiments to see how KL and entropy help to reduce Failure and Discomfort rates. For convenient comparison, we re-parametric as $\tau = w_{\cal H} + w_{\text{KL}}, \alpha = w_{\text{KL}}/\tau$ where $w_{\cal H}, w_{\text{KL}}$ control the strength of entropy and KL terms respectively. Mixing entropy and KL helps improve overall performance. To investigate the KL effect, we set the entropy coefficient $w_{\cal H}$ to $0.7$ and sweep a range of KL coefficient $w_{\text{KL}}$ to see its effects on the overall validation performance. The results are shown in Fig.~\ref{fig:KL-effect-of-SAC-ImKL}. It is noticed that a high enough value of KL helps to align the smoothness close to the reference policy ($3.84$), but too large KL causes a significant increase in the Failure metric while not improving the smoothness.
Next, we study the effect of entropy. We set the KL coefficient $w_{\text{KL}}$ to $0.5$ and sweep a range of entropy coefficients. The results are shown in Fig.~\ref{fig:entropy-effect-of-SAC-ImKL}. Our experiment shows that SAC-ImKL fails to converge when $w_{\cal H} = 0$. One can observe that a value of $0.7$ seems to drive as smoothly as BC-perturb while achieving the best overall performance. A higher entropy value (e.g. $w_{\cal H} = 4$, $w_{\cal H} = 7$), exhibits a similar and even better Failure score than $w_{\cal H} =0.7$ while significantly outperforming in the smoothness compared to BC-perturb. Therefore, we can observe that increasing the entropy coefficient is beneficial in dropping both discomfort rate and failure cases, suggesting that entropy regularization helps.

\subsubsection{In-depth collision analysis}
\label{sec:collision-analysis}

Table~\ref{tab:in-depth-collision} presents a detailed analysis of collision modes (collision front, rear, and side) on a set of 100 sampled scenarios. Note that the collision front and rear is the result of the ego planner hitting other agents, while the ego possibility does not directly cause collision side, but the planner drives in such a way that being hit by other unexpected logged agents. Regarding averaged metrics, the BC-perturb generates the highest collision rate in 4 methods. The collision rate induced by SAC-only is significantly lower than BC-perturb, however,  this method yields a high number of collision cases. SAC-ExKL outperforms -perturb in almost every collision metric except the side collision. Finally, SAC-ImKL exhibits the fewest failure rate and number of failed events, especially in the collision front and rear which the ego can actively control. All approaches have a high failure on the collision side partially because of the unexpected turn of the other agent.

\section{Conclusion}
This study shows that entropy regularization is a crucial consideration to avoid over-conservation for the RL fine-tuning process. Besides, KL-regularized RL significantly improves learning speed, human-like driving patterns, and passenger comfort. The balanced trade-off between entropy and KL-divergence shows a large improvement in safety and realism performance over baselines. Our extensive experiments examined the roles of entropy/ KL trade-off terms, RL in transformer models, and vehicle kinematic models. Along with these improvements, our work does not solve for unexpected behavior of other agents and high imitation error. Also, since our method uses IL prediction on RL data to guide the RL agent, we require a highly robust IL model (perturbation BC \cite{bansal2018chauffeurnet}) to alleviate covariate shift. Moreover, it still depends on heuristically choosing the trade-off value between entropy and KL regularization. Some considerable approaches employ offline RL taking advantage of a large amount of data available to the driver, without additional online data collection. It is perhaps treated as a pretrained approach instead of BC to mitigate distribution shifts and excessively conservative issues. 
\section*{Acknowledgment}
We acknowledge Ho Chi Minh City University of Technology (HCMUT), VNU-HCM for supporting this study.

\begin{figure}[htbp]
    \centering
    \begin{subfigure}{.5\textwidth}
    \centering
        \begin{minipage}{0.11\textwidth}    
        \end{minipage}
         \begin{minipage}{.85\textwidth}
         \centering \vspace{-3mm} \hspace{15mm} \texttt{\scriptsize{traffic elements}}
         \end{minipage}
    \end{subfigure}
     \begin{subfigure}{0.5\textwidth}
     \centering
     \begin{minipage}{0.10\textwidth}
        \centering
        \rotatebox{90}{\textbf{BC-perturb}}
    \end{minipage}
    \begin{minipage}{0.01\textwidth}
        \centering
        \rotatebox{90}{\texttt{\tiny{frame}}}
    \end{minipage}
     \begin{minipage}{.85\textwidth}
         \centering
        
        \includegraphics[width=\textwidth]{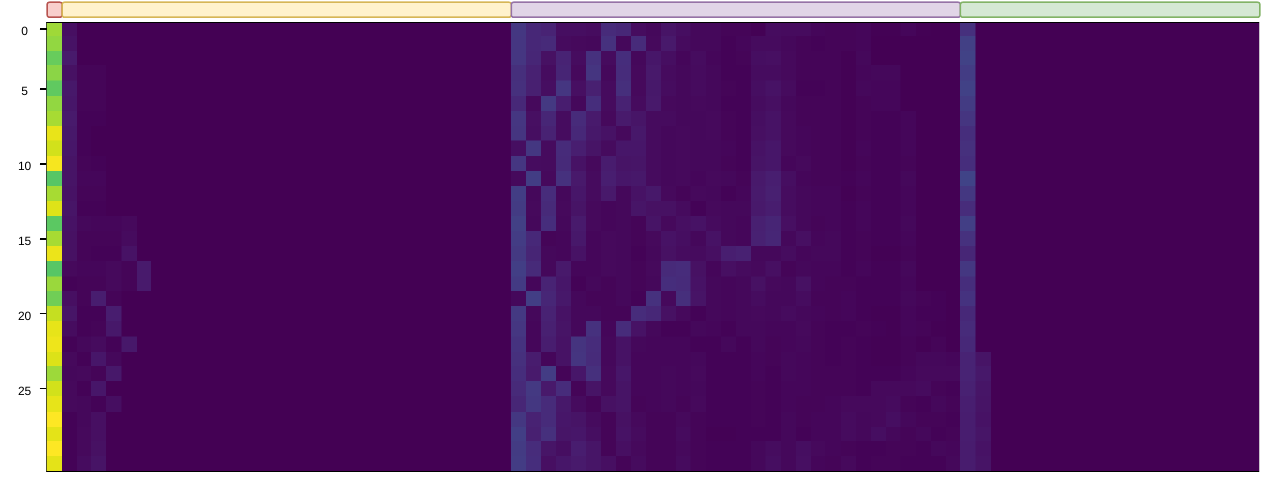}
    \end{minipage}
    \hfill
    
    \begin{minipage}{0.10\textwidth}
        \centering
        \rotatebox{90}{\textbf{SAC}}
    \end{minipage}
    \begin{minipage}{0.01\textwidth}
        \centering
        \rotatebox{90}{\texttt{\tiny{frame}}}
    \end{minipage}
     \begin{minipage}{.85\textwidth}
         \centering
         \includegraphics[width=\textwidth]{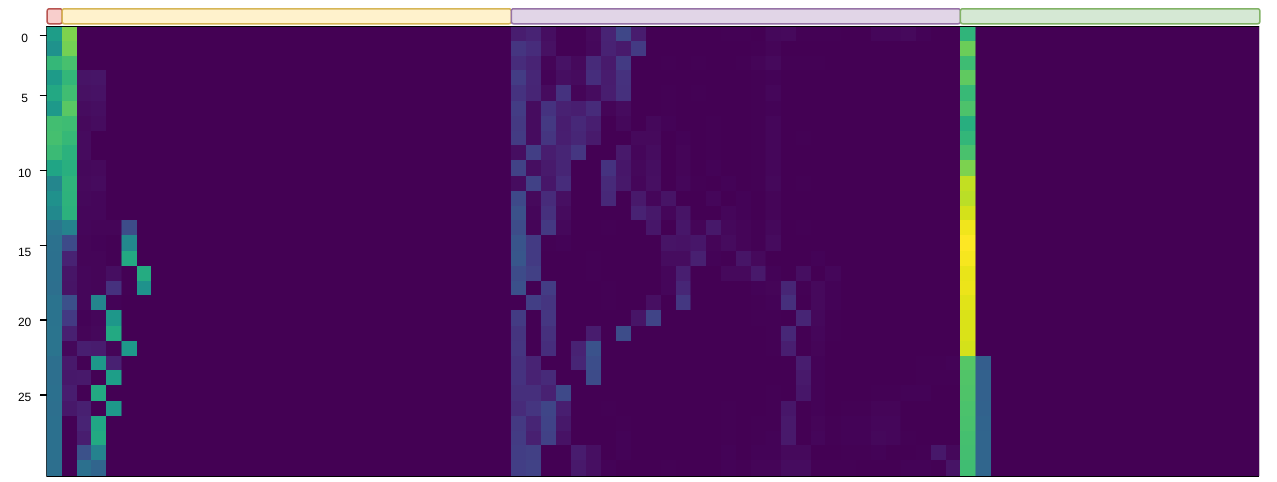}
    \end{minipage}
    \hfill

    \begin{minipage}{0.10\textwidth}
        \centering
        \rotatebox{90}{\textbf{SAC-ExKL}}
    \end{minipage}
    \begin{minipage}{0.01\textwidth}
        \centering
        \rotatebox{90}{\texttt{\tiny{frame}}}
    \end{minipage}
     \begin{minipage}{.85\textwidth}
         \centering
         \includegraphics[width=\textwidth]{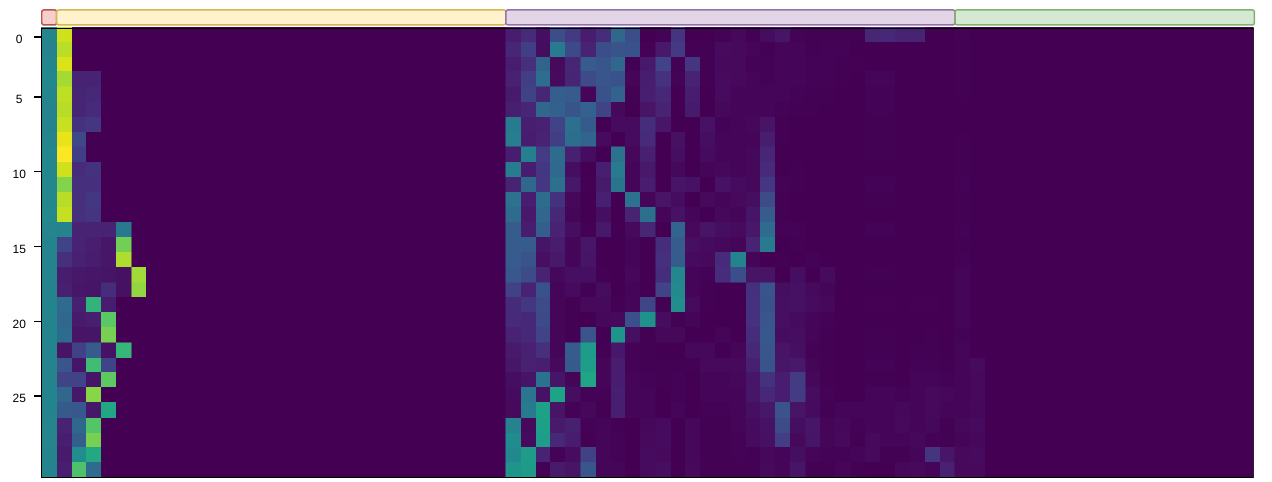}
    \end{minipage}
    \hfill

    \begin{minipage}{0.10\textwidth}
        \centering
        \rotatebox{90}{\textbf{SAC-ImKL}}
    \end{minipage}
    \begin{minipage}{0.01\textwidth}
        \centering
        \rotatebox{90}{\texttt{\tiny{frame}}}
    \end{minipage}
     \begin{minipage}{.85\textwidth}
         \centering
         \includegraphics[width=\textwidth]{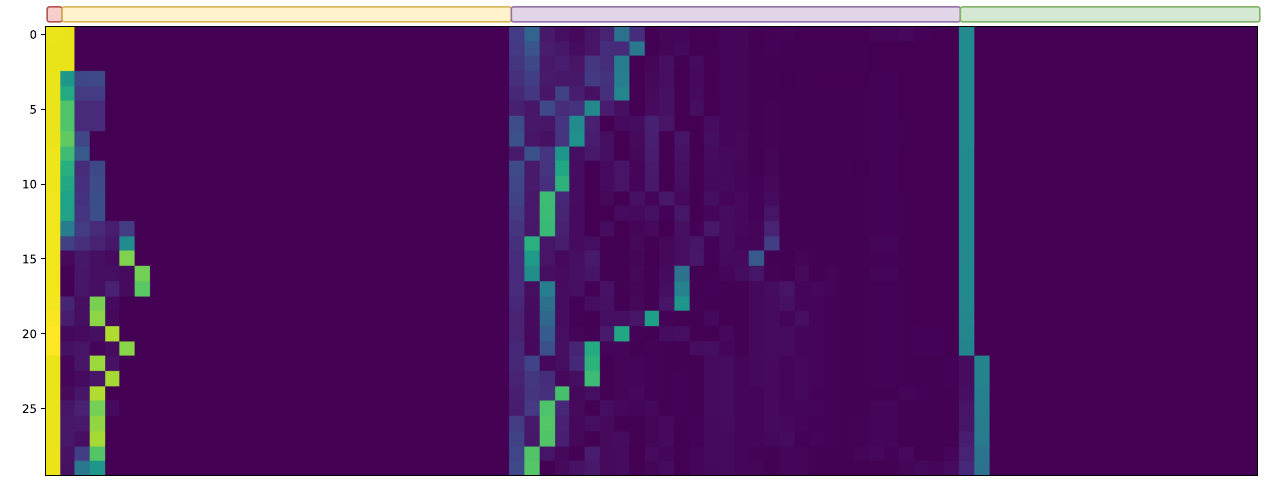}
    \end{minipage}
    \hfill
     \end{subfigure}
    \begin{subfigure}{.5\textwidth}
        \centering
        \includegraphics[width=\textwidth]{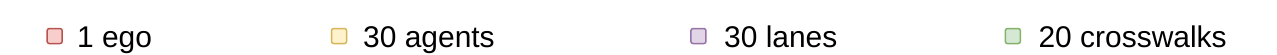}
    \end{subfigure}

     \caption{Attention heatmaps illustrate attention score between ego car and all 81 elements (agent, map) in the 30 frames of a scene.}
    \label{fig:attention_map}
\end{figure}

\appendix
\subsection{Additional Details on Model Architectures and Hyper-parameters Settings}
\label{sec:appendix_model_details}
In the training stage, we employ a common RL fine-tuning technique where the agent is first trained in a supervised manner, and then we fine-tune using the RL optimizer. We leverage the pretrained Transformer-based behavioral cloning model in paper \cite{scheel2022urban} to bootstrap our RL agent. Regarding the Transformer backbone, first, the vectorized form of the HD map extracted from the simulator is fed into a set of local encoders. The local encoder distills all points belonging to the same object into a same-size dimensional feature. These features are followed by a layer of Multi-head self-attention layer which learns the global interaction feature. The resulting features go through RL networks to predict a policy distribution and a value function. Finally, action is sampled from the policy distribution to interact with the environment and update the new state.


Since the output of BC-perturb and SAC agents are different, we simply replace the last 3 MLP layers of the IL model with the randomly initialized 3 MLP layers where the last fully connected (FC) layer outputs the desired shape of RL policy and value function. SAC also consists of two learnable parts: actor and critic networks, initialized by pretrained IL, optionally share feature extractor parameters.  In DriverGym urban driving tasks, applying 2 separate feature extractor networks results in better overall performance since it allows each network to focus on different aspects of the input data, thus leading to more accurate representations for each network.


The SAC-ExKL model is augmented with KL term $\alpha = 0.3$ while the SAC-ImKL is regularized by entropy term  $\tau = 1.2$, and KL term $\alpha = 0.4$. We use kinematic-based instead of waypoint-based action to fasten RL training. The linear learning rate schedule is used to decay from $3\times10^{-5}$ to $3\times10^{-6}$ during the 200K training step for SAC-ExKL, SAC-ImKL, and 300K training step for SAC-only, with a batch size of $256$. The discount factor is $0.8$ and the soft target update coefficient polyak $\rho$ is $0.995$. We employ the buffer capacity size of $10^5$. All experiments are implemented in RLlib \cite{liang2018rllib} framework.

\subsection{How RL benefit Transformer?}
\label{sec:rl-help-transformer}
Fig.~\ref{fig:attention_map} illustrates the attention score between ego car and all 81 elements (agent, map) in 30 frames of a scene. The indices of them are 0: ego vehicle, 1-30: other agents, 31-60: lanes, 61-80: crosswalks. IL, in the first row, heavily attends to the ego itself (recent past and current pose of the vehicle) but rarely focuses on other elements on the road. We consider it to be especially related to shortcut learning \cite{geirhos2020shortcut}, in which a deep neural network tends to find the easiest signal for driving when the context is missing. Using attention layers to understand the context by modeling relationships between elements. However, it is hard to understand the relation explicitly when the model just offline learns to imitate the expert demonstration. Otherwise, RL methods have a more intuitive attention map, which is not only aware of the ego data but also highly exploits context information from other surrounding agents, lanes, and crosswalks. This is expected because RL encourages the agent to learn how to see, how physics works, what consequences its action, etc. By combining RL with attention mechanism, we get a broader range of attention maps than BC and thus become more trustworthy in driving.


\subsection{Optimal KL reward coefficient in SAC-ExKL}
\begin{figure}[htbp]
    \centering
    \includegraphics[width=.48\textwidth]{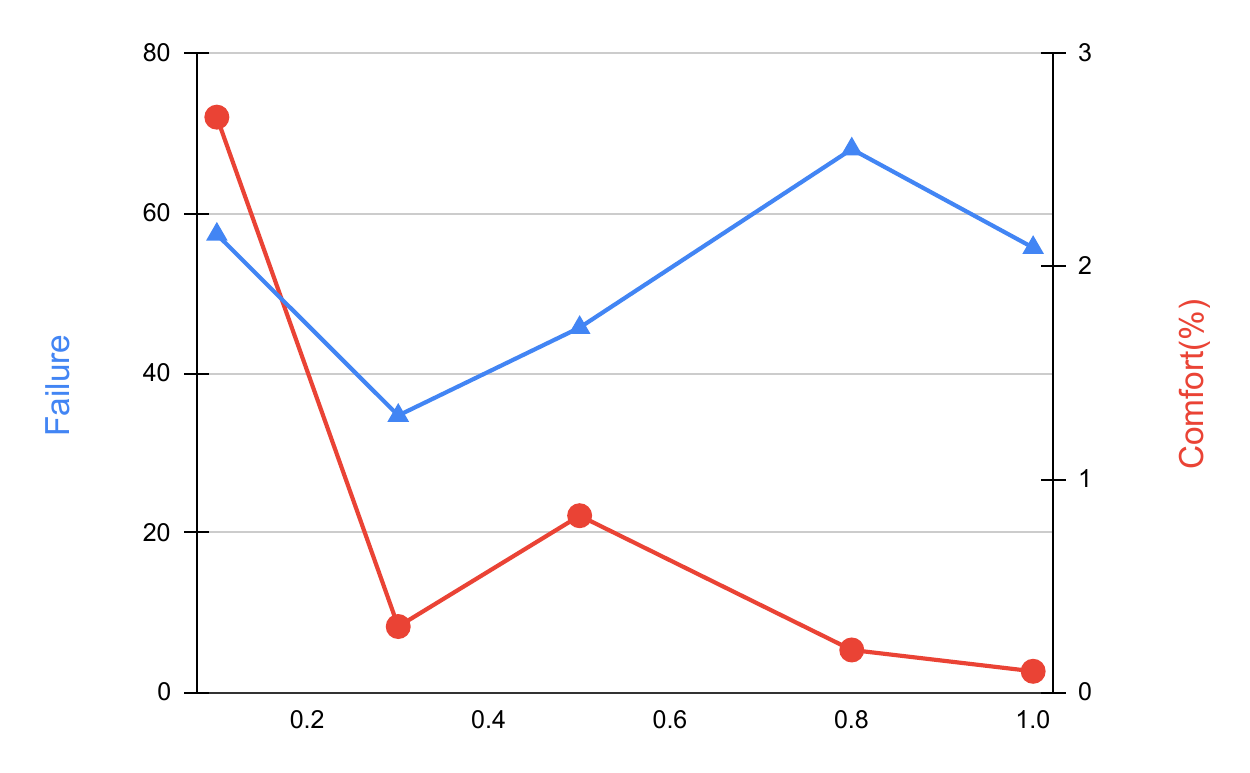}
    \caption{KL reward coefficients of SAC-ExKL.}
    \label{fig:ablation_kl_coef}
\end{figure}

In Fig.~\ref{fig:ablation_kl_coef}, we experiment with 5 different KL coefficients from 0 to 1. For easy comparison, we evaluate based on Failure and Discomfort metrics in 100 unseen scenes. The optimal value for the KL coefficient is around 0.3 when 0 results in the worst smoothness policy and 1 suffers from high Failure. Increasing KL reward results in smoother policy but suffers from poor safety and high off-road events.

\bibliography{refs}{}
\bibliographystyle{ieeetr}

\end{document}